\documentclass[conference]{IEEEtran}
\IEEEoverridecommandlockouts
\usepackage{cite}
\usepackage{amsmath,amssymb,amsfonts}
\usepackage{algorithmic}
\usepackage{graphicx}
\usepackage{textcomp}
\usepackage{xcolor}
\usepackage{colortbl}
\usepackage{booktabs} % For formal tables
\usepackage{algorithm}
\usepackage{algorithmic}
\usepackage{booktabs}
\usepackage{amsmath}
\usepackage{graphicx}
\usepackage{subfigure}
\usepackage{amsfonts,bm}
\usepackage{multirow}
\usepackage{color}
\usepackage{braket}
\usepackage{caption}
\usepackage{mathtools}
\usepackage{enumerate}
\usepackage{enumitem}
\usepackage{empheq}
\usepackage{calc}
\usepackage{dsfont}
\usepackage{url}
\usepackage{bbm}
\usepackage{bm}
\usepackage{amsthm}
\usepackage{algorithmic}
\usepackage[ruled,algo2e]{algorithm2e}
\newtheorem{definition}{Definition}

\def\BibTeX{{\rm B\kern-.05em{\sc i\kern-.025em b}\kern-.08em
    T\kern-.1667em\lower.7ex\hbox{E}\kern-.125emX}}
\begin{document}

\title{STORM-GAN: Spatio-Temporal Meta-GAN for Cross-City Estimation of Human Mobility Responses to COVID-19 \\
% {\FOOTNOTESIZE \TEXTSUPERSCRIPT{*}NOTE: SUB-TITLES ARE NOT CAPTURED IN XPLORE AND
% SHOULD NOT BE USED}
% \THANKS{IDENTIFY APPLICABLE FUNDING AGENCY HERE. IF NONE, DELETE THIS.}
}

\author{\IEEEauthorblockN{Han Bao, Xun Zhou*\thanks{* Corresponding author}}
\IEEEauthorblockA{\textit{University of Iowa} \\
% \textit{}\\
\{han-bao, xun-zhou\}@uiowa.edu
% \authornote{Corresponding author}
}
\and
% \IEEEauthorblockN{Xun Zhou*} 
% \IEEEauthorblockA{\textit{University of Iowa} \\
% % \textit{name of organization (of Aff.)}\\
% % City, Country \\
% xun-zhou@uiowa.edu}
% \and
\IEEEauthorblockN{Yiqun Xie}
\IEEEauthorblockA{\textit{University of Maryland} \\
xie@umd.edu}
\and
\IEEEauthorblockN{Yanhua Li}
\IEEEauthorblockA{\textit{Worcester Polytechnic Institute} \\
yli15@wpi.edu}
\and
\IEEEauthorblockN{Xiaowei Jia}
\IEEEauthorblockA{\textit{University of Pittsburgh} \\
xiaowei@pitt.edu}

}

% \author{Anonymous Author(s)}
\maketitle

\begin{abstract}
Human mobility estimation is crucial during the COVID-19 pandemic due to its significant guidance for policymakers to make non-pharmaceutical interventions. 
While deep learning approaches outperform conventional estimation techniques on tasks with abundant training data, the continuously evolving pandemic poses a significant challenge to solving this problem due to data nonstationarity, limited observations, and complex social contexts. %As an example, rapid developments of the pandemic situation make it challenging to gather data reflecting the most recent changes, and different regions may have different data characteristics (e.g., granularity). 
Prior works on mobility estimation either focus on a single city or lack the ability to model the spatio-temporal dependencies across cities and time periods. To address these issues, we make the first attempt to tackle the cross-city human mobility estimation problem through a deep meta-generative framework. We propose a Spatio-Temporal Meta-Generative Adversarial Network (STORM-GAN) model that estimates dynamic human mobility responses under a set of social and policy conditions related to COVID-19. Facilitated by a novel spatio-temporal task-based graph (STTG) embedding, STORM-GAN is capable of learning shared knowledge from a spatio-temporal distribution of estimation tasks and quickly adapting to new cities and time periods with limited training samples. 
% The learned initialization of the spatio-temporal cGAN can effectively and quickly be used as the start point for an unseen task. 
%In addition, we construct a novel pandemic propagation graph (PPG) to represent the spatio-temporal relationships among cities. 
The STTG embedding component is designed to capture the similarities among cities to mitigate cross-task heterogeneity.
Experimental results on real-world data show that the proposed approach can greatly improve estimation performance and outperform baselines.
\end{abstract}


\section{Introduction}\label{sec:introduction}

Evolving developments (e.g., spread, mutation, vaccination) around the COVID-19 pandemic have continued to pressure policymakers to come up with effective and changing policies that can protect public health while avoiding breakdowns of economics, and maintain the support of essential needs in daily lives. %\yl{The footnote below should be KDD'22, check the paper format.} 
To mitigate this dilemma, staged reopening to avoid infections caused by eased social distancing policies have been implemented. As a result, estimation on dynamic human mobility responses to the pandemic condition and policies remains a crucial task in policymaking.
% for policymakers who aim to reduce infections and protect the economic system. 

Due to the asynchronous spread of the disease, it is particularly hard for cities in the early stage of an outbreak or wave (e.g., the Omicron variant) to estimate future human mobility responses under unprecedented severity levels or unseen policies as there is very limited historical data. To address this issue, it is crucial for such cities to be able to leverage other cities' past experiences and knowledge for its own estimation. To this end, mobility response estimation methods that can leverage cross-city knowledge to achieve promising results are urgently needed. 

%In this work, we measure human mobility responses by the number of visits to points-of-interest (POIs) such as grocery and hardware stores, restaurants, gas stations, etc.,
% Estimation of such human mobility responses is challenging due to the complex and sometimes unknown social contexts, as well as limited training data. Moreover, the continued and drastic changes of the pandemic further limits the availability of recent or up-to-date data that can reflect such dynamic changes, which is especially challenging to leverage the approximation power of data-driven deep learning approaches.
% Responding to the urgent need by policymakers and public health experts, w
In this paper, we make the first attempt to solve the {\em cross-city human mobility responses estimation problem}: Given a set of inputs on contextual (e.g., population, point-of-interest {\em POI} counts), epidemic (e.g., COVID-19 cases), policy (e.g., stay-at-home orders) conditions and corresponding human mobility responses measures (e.g., POI visit counts, home dwell time) from multiple cities, we aim to learn a model, which can quickly adapt to previously unseen cities and time periods, and estimate human mobility response dynamics under any projected conditions.%\footnote{In this work, we measure human mobility responses by the daily visit counts to points-of-interest (POIs) such as grocery and hardware stores, restaurants, gas stations, etc. However, other measures can be used with our method without any problem.}.

%using shared-knowledge extracted from a distribution of estimation tasks sampled from multiple cities and timestamps. The shared-knowledge will be leveraged to quickly adapt to new estimation tasks from a new city. 
% Note, the data from the target city is not contained in the meta training set, and here the input conditions might not have been observed in the targeted area in the historical data. 

\noindent{\bf Challenges.} 
The cross-city human mobility response estimation problem has three major challenges. 
\emph{First}, human mobility responses depend on many complex social-physical factors which are unknown or uncertain. For example, responses can be affected by people's willingness in cooperating with policies, decisions from service providers (e.g, whether a restaurant will open or allow dine-in options), changes in public transportation, supply, and many more \cite{kraemer2020effect}.
\emph{Second}, human mobility responses often have spatio-temporal non-stationarity. For example, the contribution of different factors in mobility tends to vary from region to region due to cultural and economic differences, and can quickly evolve over time.
Such spatial and temporal nonstationarity greatly limits the availability of training data for each estimation task, making it difficult to leverage the approximation power of data-driven approaches.
\emph{Third}, there also exist complex spatial and temporal dependencies across different estimation tasks, i.e., cities and time periods, which need to be explicitly considered for robust parameter sharing. For example, cities may share similar mobility dynamic patterns based on their spatial adjacency (e.g., distances, travel connections such as airlines), and their stages in the pandemic. 
%Note that the stages may not be reflected directly by the timestamp, and are impacted by other factors such as the start time of disease spread. Thus, the learning framework needs to be flexible in representing such spatio-temporal relationships.

\noindent{\bf Related Work.} 
% Many recent efforts have attempted to understand the impact of human mobility in the COVID-19 transmission \cite{soucy2020estimating,kraemer2020effect}. While these studies have demonstrated the importance of human mobility responses, they do not address the challenges in mobility estimation or simulation. 
Many recent efforts have attempted to use machine learning methods for spatio-temporal estimation tasks. For example, \cite{zhang2019trafficgan} uses a conditional generative adversarial network (cGAN) to estimate traffic volume. Similarly, a recent work \cite{10.1145/3397536.3422261} proposed a COVID-GAN for human mobility estimation, where policies (e.g., school closure) are used as constraints to help improve estimation results. However, both of them only consider estimation in a single city, without modeling the spatial and temporal dependencies across cities or stages. Therefore, they lack the ability to quickly adapt to unseen cities. In addition, COVID-GAN is a purely spatial model, which does not explicitly model the dynamics of human mobility responses over time. A POI embedding transfer learning approach is proposed \cite{jiang2021transfer} to predict urban traffic from one city to another city. This approach adapts model parameters without using meta-learning method. 
% although its share data \cite{} spans across multiple weeks, time is only modeled as a feature (i.e. week ID), and its network architecture adopts a snapshot-based approach
% does not consider the temporal correlation, and the model structure does not consider the influence of nearby environments when estimating the human mobility for each sample. 
%
In recent years, many meta-learning approaches have been proposed to solve the few-shot learning problem~\cite{liang2020dawson,yao2019learning}, where the training samples are limited for new tasks. However, few of them are designed for spatio-temporal tasks. Among the exceptions, \cite{zhang2020cst} uses a model-based meta-learning approach with a variational autoencoder structure to generate traffic volume. Another traffic prediction work \cite{yao2019learning} combines the attention mechanism and CNNs, and uses functionality zones to group cities into tasks before applying the model-agnostic meta-learning (MAML \cite{finn2018probabilistic}) framework.
%\yl{Add an ref for MAML}. 
However, this model is designed with no time-based tasks, which is insufficient to model the continued and dynamic changes in our problem. Moreover, it does not model and utilize the spatio-temporal dependencies and correlations across different tasks.
%Meta-learning has been studied as an powerful solution to address the data scarcity problem, the ability of learning quickly and effectively that utilize prior knowledge and experiences provides make it become a population method in recent machine learning domain. The assumption of meta-learning is tasks are implicitly related which allow the meta-learning methods to learn a shared parameters. If the underlying data distribution are significantly different between tasks, the shared knowledge transfer will make no contribution. 

\noindent{\bf Proposed Work.} 
To address the limitations of prior works, we formulate the problem as a meta-learning-based conditional data generation problem, where each task of the meta-learning framework is to estimate a time-series of human mobility response maps for a specific city during a specific time period under designated conditions. 
As a solution, we design 
a \underline{S}patio-\underline{T}emp\underline{OR}al (conditional) \underline{M}eta-\underline{G}enerative \underline{A}dversarial \underline{N}etwork (STORM-GAN). Building on top of a conditional GAN (cGAN) model~\cite{jiang2020covid}, the STORM-GAN model learns to generate spatio-temporal mobility dynamics in different cities under a set of geographic, epidemic, social and other factors. It utilizes a meta-learning paradigm to learn a general model initialization from a distribution of tasks (i.e., mobility estimation for each city over a time period) for fast adaption to new spatio-tempral tasks (e.g., new cities, future projection). To explicitly model the spatio-temporal relationships across tasks, we propose a \underline{S}patio-\underline{T}emporal \underline{T}ask-based \underline{G}raph (STTG) embedding method for better model generalization and adaptation, which further improves STORM-GAN's performance.
%
%The use of a conditional GAN (cGAN) as the base model allows the consideration of known and uncertain factors \cite{gauthier2014conditional}.  %The ST-cGAN framework consists of a CNN and an LSTM structure which jointly capture the spatio-temporal correlations. 
%We incorporate LSTM and CNN layers into a cGAN to explicitly capture spatial (within-task) and temporal mobility patterns. Moreover, STORM-GAN operates on a distribution of spatio-temporal tasks constructed from time-sequences in multiple cities, and utilizes a meta-learning %variant of the model-agnostic meta-learning (MAML) 
%
% we adopt the meta-learning problem setting to solve the third challenge by learning a well-generalized initialization from multiple cities for adaptation. The initialization can be easily adapted to unseen tasks sampled from the unseen city.
% To further address the spatio-temporal heterogeneity, 
%Finally, to explicitly model spatio-temporal relationships across tasks, we propose a COVID-19 spreading graph to learn spatio-temporal embeddings of cities and measure task similarities for better generalization and adaptation of the meta-model.
%
% embedded to enhance the task-wise learned parameters and help to measure the task similarities.
Overall, the {\bf contributions} of this paper are as follows:
%\vspace{-40pt}
\begin{itemize}[leftmargin=*]
    \item We formulate the cross-city human mobility response estimation problem as a spatio-temporal meta-learning-based data generation problem. To the best of our knowledge, this is the first attempt to estimate human mobility through a deep meta-generative framework. 
    % from the meta-learning perspective across cities and timestamps.
    % propose a spatio-temporal cGAN model (ST-cGAN-Meta). The proposed model utilizes the MAML \cite{} optimization-based meta-learning framework to update meta-learner, which is trained on a distribution of estimation tasks sampled from multiple cities with the goal of learning a high-level representation that can quickly be adapted to related unseen tasks from a new city.
    The proposed novel meta-generative framework models the uncertainty, spatial and temporal patterns simultaneously. %The MAML-based meta-learner enables fast adaption across tasks from multiple cities and timestamps. spatio-temporal correlation. Specially, 
    \item Specifically, we propose a COVID-19 spatio-temporal task-based graph, which is embedded into the framework %via a graph convolutional network 
    to explicitly model spatio-temporal dependency among different tasks, further improving the learning of the shared-knowledge.
    % capability to learn a shared knowledge and improve solution quality.
        % that considers uncertainty \cite{}, spatio-temporal dependency (e.g., a distribution of tasks based on cities and timestamps), and spatio-temporal relationships across tasks (e.g,, by travel connections, pandemic stages, etc.).
    \item We perform various experiments on real-world datasets 
    %from 6 cities spanning over 35 weeks 
    to evaluate the performance of the proposed approach under different scenarios, and the results show that STORM-GAN can greatly improve mobility response estimation compared to other candidate approaches. We have released our code and the sample data in a temporary GitHub link.\footnote{https://github.com/BaoHan88/STROM-GAN.git}

    % \item We made our code and sample data available to contribute to the research community.
    % \footnote{https://www.dropbox.com/sh/ivr6ouf5t3rjr22/AAAilRhuM94sNFA1OhjB
    % UhPXa?dl=0}
    
    % We made our code and sample data available to contribute to the research community}\footnote{\url{https://www.dropbox.com/sh/mtn7agidbo4flw4/AADv5P_CSgp7T_f-WuET5fv4a?dl=0}}.
\end{itemize}

\section{Problem Statement}\label{sec:problem}
\begin{figure*}
  \centering
  \includegraphics[scale=0.51]{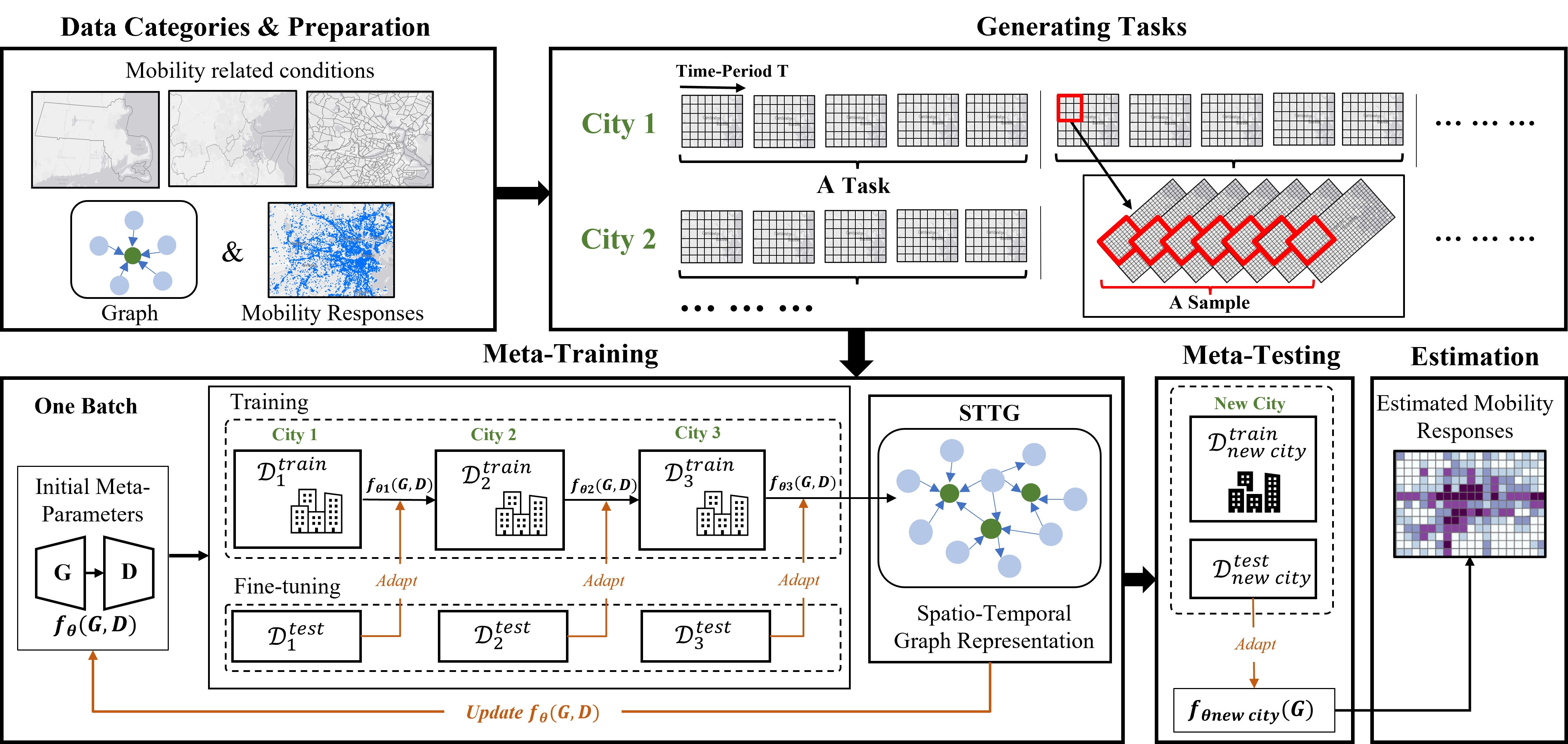}
  \caption{Overall framework.\vspace{-5pt}
  }
  \label{fig:framework}
\end{figure*}

This section introduces a set of basic concepts about our data modeling, and then provides a formal problem statement. The overall solution framework is shown in Fig. \ref{fig:framework}. %, and then introduces the solution framework on meta-learning adaptation. 
%Table 1 lists the notations used in the paper. 
% \usepackage{booktabs}
% \usepackage{graphicx}
% \begin{table}[]
% \centering
% \caption{Notations}
% \label{tab:notation}
% % \small
% % \resizebox{\columnwidth}{!}{%
% \begin{tabular}{|p{1.5cm}|p{5.5cm}|}
% \hline
% Notations    & Descriptions                              \\ \hline
% $s_{i}\in S$   & A spatial region corresponding to a $l\times l$ window in a spatial grid $S$                   \\ \hline
% $t \in T$       & A slot or timestamp in a time-period  \\ \hline
% $\textbf{M}^{t}$        & Human mobility response at ${t}$          \\ \hline
% $\textbf{F}^{t}$        & Mobility related features used to estimate ${\textbf{M}^t}$        \\ \hline
% $N$         & Number of tasks                           \\ \hline
% $k$  & Number of conditions \\ \hline
% $\mathcal {T}_i$ & An estimation task                       \\ \hline
% $\theta$       & Meta-learner parameters                   \\ \hline
% $\theta_i'$         & Task-specific parameters for $\mathcal {T}_i$ used in meta-training \\ \hline
% \end{tabular}%
% % }
% \end{table}

\subsection{Basic Concepts}\label{sec:concept}
\begin{definition}
{\textbf{Spatial grid}} $\textbf{S}$ is a grid-discretization of a spatial field (e.g., a city), where each grid cell $\textbf{s}_i$ represents an equally-sized squared area. Given $\textbf{S}$, the location of any POI can be mapped into a grid cell. For simplicity, in this work we choose the grid cells to be $1km\times 1km$.
\end{definition}

\begin {definition}
{\textbf{Temporal period}}
$\textbf{T}$ is a temporal period (e.g., a 7-day window) containing equal-length slots (e.g., a day), denoted as $T = \left \{ t_{1}, t_{2}, ..., t_{n}\right \}$, where each slot $t$ represents the finest temporal resolution of the data.
\end{definition}

\begin{definition}\label{def:conditions}
{\textbf{Mobility related conditions:}}
All conditions that will influence human mobility responses are mobility related conditions including contextual conditions (e.g., population, household income), epidemic conditions (e.g., COVID-19 confirmed cases and deaths), and policy conditions (e.g., strict stay-at-home or shelter-in-place orders). We denote a list of $\textbf{k}$ conditions as $\textbf{F} = \left \{ f_{1}, f_{2}, ..., f_{k}\right \}$. For a grid cell ${s}$, we denote $f^{s,t}$ as all the conditions of ${s}$ in time slot ${t}$.
\end{definition}

\begin{definition}\label{def:reponses}
{\textbf{Human mobility responses:}}
The human mobility responses $\textbf{M}$ is a two-dimensional tensor, representing the total number of visits to POIs (e.g., grocery stores, hardware stores, restaurants, gas stations) in each grid cell ${s}$ for time slot ${t}$. 
\end{definition}
Note here we use POI visit counts simply as an example to demonstrate the solution framework. Other mobility measures 
(e.g., median home dwell time) can also be used with our model. The choice of the measure is not the focus of this paper.

\begin{definition}
{\textbf{Generator $\textbf{G}$:}}
A deep neural network model that is used to generate a series of human mobility response maps $\textbf{M}'_G$ given a set of conditions.
%Model that is used to generate new plausible examples from the problem domain.
\end{definition}

\begin{definition}
{\textbf{Discriminator $\textbf{D}$:}}
A deep neural network model that outputs a probability $p_{real}$ at which a map of human mobility responses is classified as from real-world rather than from a generator ${G}$.
%Model that is used to classify examples as real (from the domain) or fake (generated)
\end{definition}

\begin{definition}\label{def:sttask}
{\textbf{Spatio-temporal mobility estimation tasks:}}
A task $\mathit{\mathcal{T}_i}$ consists of a series of pairs $\left ( \mathbf{M}^{t}, \mathbf{F}^{t}  \right )$ for a few consecutive time periods ${T}$ (e.g., 5 weeks) in a partitioned area $\textbf{S}$ (e.g., $10 \times 10$ grids of a city). Each $\textbf{sample}$ is a 4D tensor with size $l \times l \times k \times |T|$, where $l \times l$ is the size of the spatial window, and ${k}$ is the number of conditions. Each $\mathit{\mathcal{T}_i}$ is divided into a training set $\mathit{\mathcal{D}_{i}^{train}}$ and a testing set $\mathit{\mathcal{D}_{i}^{test}}$.

% The training set is used to learn temporary task-specific weights (i.e., in the inner loop of MAML), and the test set is used for meta-weight update (i.e., outer loop).
\end{definition}

\subsection{Problem Definition}\label{sec:metapd}
We construct the tasks (Def. \ref{def:sttask}) by a spatio-temporal partition of all conditions $\textbf{F}$ and mobility responses $\textbf{M}$. Each spatio-temporal task $\mathcal{T}_i$ contains data from the grid $S$ of a single city for $w$ consecutive time periods $\{T_1, ..., T_w\}$, and tasks are mutually exclusively (i.e., no overlap along the temporal dimension). Each data sample in a task contains a time-series of length $|T|$ with any start time (but the time-span of a sample must be completely within the span of a task).  %The general problem is defined as:
% For simplicity, the following formulation shows the inputs and output for a single spatio-temporal task $\mathcal{T}_i$, and the meta-learning framework on top of it will be discussed in Sec. \ref{sec:maml}.

\noindent\textbf{Inputs:}
\begin{itemize}[leftmargin=*]
    % \item A grid partition $S_c$ of each city;
    % \item Conditions (features) $\{\textbf{F}^{t-|T|+1}_c,..., \textbf{F}_c^t\}$ and mobility response $\textbf{M}_c^t$ for each $S_c$;
    \item A time-series of conditions  $\{\textbf{F}^{t-|T|+1},..., \textbf{F}^t\}$ for each data sample in training tasks;
    \item Mobility response $\textbf{M}^t$ for each sample in training tasks;
    % \item $\{\textbf{F}_{test}^{t-|T|+1},..., \textbf{F}_{test}^t\}$ for test samples (from a new task).
\end{itemize}
\textbf{Outputs:} 
\begin{itemize}[leftmargin=*]
    \item A generator $\textbf{G}$ to generate/estimate mobility responses;
    \item A meta-initialization $\theta$ for $\textbf{G}$ for fast adaptation to training and new testing tasks.
    % \item Estimation of mobility responses $\textbf{M}_G$;
\end{itemize}
\textbf{Objective:}
\begin{itemize}[leftmargin=*]
    \item Minimize average generation error on new testing tasks. 
\end{itemize}

In this work, given a series of tasks sampled from multiple cities (e.g., Boston and NYC), we train a meta-generative model. When a new city (e.g., Houston) comes in with a small training set, we quickly fine-tune the meta-model parameters to obtain a tailored model for the new city to generate its mobility responses.

%Note that the conditions $\mathbf{F}^{t}$ (i.e., mobility related features) for $\mathbf{M}^{t}$ is not from the same time-slot but the previous one for two reasons: (1) We expect the generator to estimate mobility responses for the next timestamp rather than the 'present' (i.e., a projection fashion); and (2) There may be a delay in mobility responses to the conditions. Each sample is a time-series of length $|T|$. Thus, the model is able to observe a longer history of conditions and is not limited to a snapshot with a one-timestamp delay. 

\section{Methodology}\label{sec:method}
\begin{figure}
  \centering
  \includegraphics[scale=0.555]{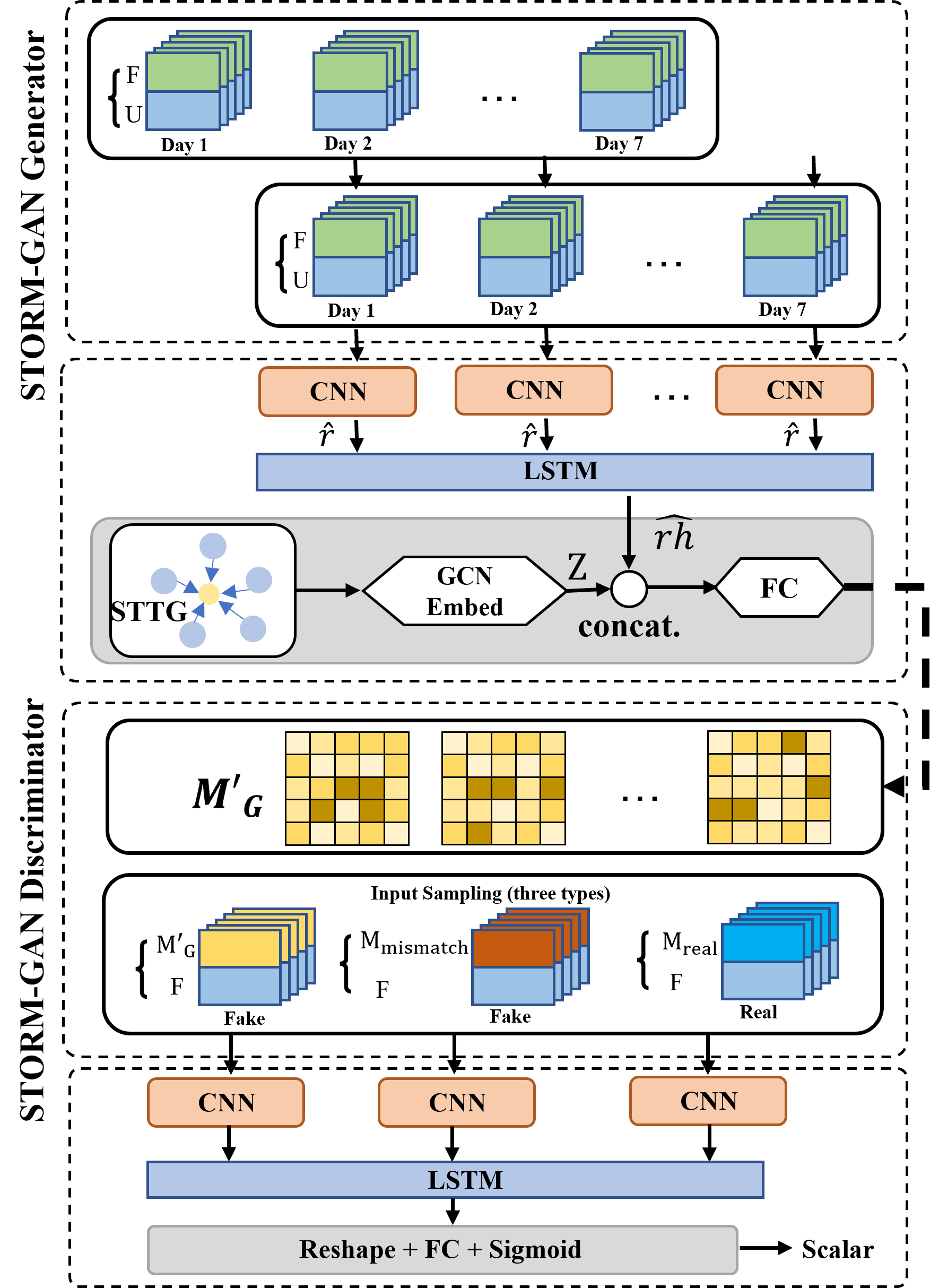}
%   \vspace{-5pt}
  \caption{STORM-GAN architecture.
  \vspace{-15pt}
  }
  \label{fig:framework2}
\end{figure}

In this section, we present the details of STORM-GAN. We first present the architecture of the proposed spatio-temporal meta-generative model. Then, we show the details of meta-parameter updates in STROM-GAN. The model architecture is shown in Fig. \ref{fig:framework2}.
%\xz{need to describe fig. 2 briefly}

% We first present a spatio-temporal GAN architecture with CNN and LSTM. Then, we introduce an pandemic propagation graph (PPG) to help enhance the learning of the cross-task initialization in the meta-learning based STORM-GAN. Finally, we show the details on meta-parameter updates in STORM-GAN.

\subsection{STORM-GAN with Spatio-Temporal Task-based graph-embedding.}
We now provide details for the three components in our proposed spatio-temporal meta-generative model: spatio-temporal generator, discriminator and spatio-temporal task-based graph embedding. 
% As discussed in Intrudiction section, both spatial and temporal patterns are important for characterizing human mobility responses. To capture such complex patterns, we propose a spatio-temporal generative network. which is a conditional GAN that is composed of a spatio-temporal generator ${G}$ and a discriminator ${D}$. %; CNN and LSTM are sub-components of the generator. 
% Furthermore, to help improve the estimation quality, we propose a pandemic propagation graph (PPG) and
% % to measure similarity among spatio-temporal tasks (Def. \ref{def:sttask}), and 
% learn embeddings of the subgraphs to represent underlying spatio-temporal relationships among tasks using a graph convolutional network. In the following, we will discuss the three key components in the network architecture. %, i.e., the spatio-temporal generator, discriminator and PPG-based graph embedding. 

\subsubsection{Spatio-Temporal Generator}
The spatio-temporal generator aims to generate human mobility responses while capturing the spatial patterns and temporal dependencies. The utilization of GAN structure allows known factors be learned as conditions and unknown factors be represented by latent noise which help the model to express these uncertainties (i.e., mobility response estimations may have some degree of variations).
As shown in Fig. \ref{fig:framework2}, the generator uses a stack of CNN and LSTM elements where CNN captures local spatial patterns and maintains the spatial representation (e.g., neighbor relationships), and LSTM is able to capture temporal trends in a given sequence. The generator takes a condition tensor $\textbf{F}\in \mathbb{R}^{l\times l\times k\times |T|}$ (we skip the batch dimension here for simplicity) 
and a latent code tensor $\textbf{U}\in \mathbb{R}^{l\times l\times u\times |T|}$, where $k$ is the number of conditions (e.g., policy, COVID statistics and contextual conditions), 
$u$ is the dimension of the noise vector for modeling the uncertainties, and $|T|$ is the length of a time period. 
%The other part of the input tensor $\textbf{K}$ contains the domain knowledge based constraints that will be used in the second phase of the generator (Sec. \ref{sec:dk}). $l \times l$ is the size of the spatial window of a data sample,
% we will introduce in Sec. \ref{sec:evalution}, 

In the generator, %the input tensor first goes through three convolutional layers, where each convolutional layer uses a $3\times 3 \times (k+u)\times 128$ kernel with zero padding ($\textbf{F}$ and $\textbf{U}$ are concatenated together), followed by batch normalization and ReLU activation; where 128 is the number of output features for CNN layers.
%After the convolutional layers, we apply a flatten operation to vectorize the output tensor, and reduce the dimension through a fully connected layer. 
% Now, the learned feature representation denotes as $\hat{r}_{{s}_i}^t$ maintains the spatial correlation for each time period $t$ of region ${s}_i$. 
denote the CNN output as $\hat{r} \in \mathbb{R}^{d\times |T|}$, where $d$ is the number of output features.
% $s_i$ represents the spatial region corresponding to the input $l\times l$ window in grid $S$, and $t$ is the last (most recent) timestamp of the period, and 
% _{{s}_i}^t
% LSTM can learn sequential correlations by containing memories from previous timestamps. 
Next, to capture temporal patterns and trends, $\hat{r}$ is fed into a LSTM layer, where the memory vector is concatenated to $\hat{r}$.
% Since we aim to generate a series of human mobility responses maps, we adopt LSTM to model the temporal trends by maintaining a memory vector through the training. The output $\hat{r}_{{s}_i}^t$ from CNN is concatenated with the memory vector as the input of the LSTM layer. 
Then, the output from the last timestamp of the LSTM layer $\hat{rh}$ will be concatenated with the graph embedding, and further passes through a fully connected layer to generate the final output.
% $\textbf{M}_G$
% , i.e., estimated human mobility responses.
% $\hat{\textbf{M}}_{{s}_i}^t$.
% maintains both the spatial and temporal relations for region ${s}_t$ during time period $t$. 
This $\hat{rh}$ is not yet the estimated mobility response $\textbf {M}'_G$.
% Before generating $\textbf{M}_G$, 
For more robust estimation, the spatio-temporal generator additionally uses a proposed spatio-temporal task-based graph embedding to characterize task-level spatio-temporal features and potential dependency across multiple cities and their mobility patterns, as discussed in the next section.
% Before feed the estimated human mobility $\textbf{M}_G$ from generator into the discriminator to determine whether the generated value is from ground truth or produced by generator, we enforced a pandemic propagation graph embedding %$\textbf{E}$ 
% to co-train with the network architecture.           
%
\subsubsection{Spatio-Temporal Task-based Graph (STTG) Embedding}\label{sec:ppg}
%\xz{Double check this part.} 
In real-world scenarios, spatial meta-learning tasks may have a very diverse distribution. For example, in our problem, tasks sampled from multiple cities can have significantly different human mobility patterns due to different urban contexts. Meanwhile, there may also exist underlying dependencies among cities due to traffic connections, geo-socio similarities, etc. Such spatial distribution of tasks, if properly utilized, would greatly enhance the performance of the learned meta-learning model.
%For example, cities of similar sizes, socio-economic environment, and in similar stages of the pandemic may share similar human mobility pattern that could help the estimation of new cities. 
%For example, mutual travel connections (e.g., flights) and stages of pandemic may impact the similarity between different tasks.
% that poses another challenge for learning the meta-level shared knowledge cross all tasks.     
%  including mobility responses during COVID-19
% Recently, external features assisted learning has becoming an emerging trend, which can help reduce learning difficulty and improve learning results.
% The external features that influence the mobility pattern during COVID-19 pandemic range from observations to social and physical theories. Inspired by recent research \cite{} in exploring the relationship of human mobility and COVID-19 , 

\indent To better model heterogeneity and dependency across spatio-temporal tasks, we propose a novel spatio-temporal task-based graph ({STTG}) to incorporate such information and facilitate the learning of transferable knowledge from related tasks. In the following part, we will first introduce the construction rules of {STTG}, and then discuss STTG-based embedding learning.   

The STTG in our proposed STORM-GAN framework is a directed weighted graph $\mathcal{G}\left ( \mathcal{V},\mathcal{E}  \right )$, where nodes represent the spatial locations of tasks (e.g., cities) and edges (and weights) represent the relevance among spatial locations. The graph is attributed, meaning that the nodes are associated with attributes $f(\nu_i)$ to describe the characteristics of each spatial location in the task space.

STTG can be defined in various ways depending on the underlying analysis goal and the network data used. In our particular application, we define each node $\nu_i$ as a major metropolitan area in the U.S, which contains features $f(\nu_i)$ of the city such as the current stage of the pandemic. Each edge $e_{ij}$ connecting cities $\nu_i$ and $\nu_j$ indicates that there is geo-socio similarity between $\nu_i$ and $\nu_j$ in the pandemic, where the edge weight represents the strengths of such similarity. Depending on how ``similarity'' is measured, we can define the edge and weights differently. Examples of such measures may include the infection spreading between cities \cite{kang2020multiscale}, correlation between cities' mobility patterns, etc.

In this paper, we present two examples of STTG construction cases, although other definitions can be used with our method as well. In the first case, we define the edges and their weights based on physical reachability, i.e., the number of direct flights and driving distance between cities, with the assumptions that the COVID spreading is tightly related to traveling and that cities with stronger transportation connections tend to have more relevance in COVID situation. In the second case, we define the edges based on the similarity of historical mobility pattern distribution measured by the Kullback–Leibler (KL) divergence\cite{kullback1997information} between cities. We provide details on the STTG construction in Sec. \ref{sec:sttg}, and show effectiveness in Sec. \ref{sec:evaluation results}.

Next, we use the built STTG in the meta-training phase to help learn more useful knowledge across tasks. As Fig. \ref{fig:framework2} shows, during the training on generator, we first sample a task-specific 1-hop subgraph $\textbf{H}$ for the corresponding node (a city) on the STTG. Then, we obtain a sub-graph embedding using Variational Graph Autoencoder (VGAE) which consists of Graph Convolution Neural network (GCNs) \cite{chen2020simple} by solving:
\begin{equation}\label{eq:gcn}
%\small
    f(\textbf{X}^{L}, A) = \alpha(\hat D ^{-\frac{1}{2}} \hat A \hat D ^{-\frac{1}{2}}\textbf{X}^{L-1}W^{L-1}),
\end{equation}

\noindent where $A$ is the adjacency matrix, ${\hat A = A + I}$, ${I}$ is the identity matrix, ${\hat D}$ is the diagonal node degree matrix of ${\hat A}$, $\alpha(\cdot)$ is a activation function (e.g., ReLU), $X$ is the feature matrix of each node from the graph, and ${W}^{L-1}$ is a weight matrix for the ${L-1}^{th}$ layer. The encoder takes $A$ and $X$ as inputs and generates the latent variable $\mathcal {Z}$ as output. The decoder reconstructs a adjacency matrix defined by the inner product between latent variable $\mathcal {Z}$.

The graph feature representation $\mathcal {Z}$ is concatenated with the output $\hat{rh}$ (Fig. \ref{fig:framework2}), and flows through a final fully connected layer in the spatio-temporal generator to achieve $\textbf {M}'_G$.
The new STTG and GCN-based embedding, being part of the generation process, will also help the meta-learner to incorporate the similarity and dependency among tasks. 
% In addition, the construction of PPG can be generalized by other network data that represent the infection spreading between cities \cite{kang2020multiscale}, the proposed idea provides a data-agnostic framework to help the task relatedness learning.% during meta-training.
% . This graph representation will help to learn the task relatedness and help the meta-learning model to capture spatial-temporal correlations that underlying in tasks. 
% _{{s}_i}^t

\subsubsection{Spatio-Temporal Discriminator}\label{sec:dis}
Fig. \ref{fig:framework2} shows the structure of the discriminator, which takes a tensor of size $\mathbb{R}^{l\times l\times (k+1)\times |T|}$, where $k$ is the number of conditions (same as that for generator) and the added one dimension is for the mobility response layer.

% The mobility response layer is either the generated $\textbf{M}_G$ or real $\textbf{M}_{real}$. 
% For the discriminator, samples with "fake" labels are created by: (1) combining $\textbf{M}_G$ with its corresponding features (conditions) used for the generation, or (2) combining $\textbf{M}_{real}$ with an unrelated feature tensor (e.g., $\textbf{F}$ from a different $l\times l$ window or time range). Samples with "real" labels can only be created by combining $\textbf{M}_{real}$ with its own features.
% Using these inputs, the discriminator learns to determine whether an input is "real" or "fake".
% % The discriminator utilizes the ground truth mobility data $\textbf{M}_{real}$ to judge whether the generated image $\textbf{M}_G$ is a real or fake one.  
% %  outputs a generated sequence of human mobility maps as input for Discriminator. 

To create training data for ``fake" or ``real" labels, the input tensors are created in three ways: (1) generated mobility $\textbf {M}'_G$ concatenated with conditions; (2) real mobility $\textbf{M}_{real}$ concatenated with corresponding conditions; (3) conditions concatenated with mismatched real mobility $\textbf{M}_{mismatch}$. Only samples from the second combination are labeled ``real". Using these inputs, the discriminator learns to determine whether an input is ``real" or ``fake". 

%Similar to the generator, the discriminator contains three CNN layers (kernels $\in \mathbb{R}^{3\times3\times(k+1)\times 128}$) with batch normalization and ReLU activation. The output of last CNN layer is fed into a LSTM layer, it is activated by ReLU and followed by batch normalizations. The output of LSTM is reshaped into a vector and passes through a fully connected (FC) layer, with a Sigmoid activation, and outputs a final scalar, indicating the guess on whether the input sample is "real".
STORM-GAN training on a single city 
is performed through adversarial configuration between the generator and discriminator. A min-max objective function is used to train ${G}$ and ${D}$ jointly by solving:
\vspace{-10pt}
\begin{equation}\label{eq:obj}
\small
\begin{split}
    \mathcal{L}_{G,D} = %\min_G \max_D V(G, D) = 
    \text{E}_{\textbf{M}\sim P_{data}} [\log D(\textbf{M},\textbf{F})]\\
    +\, \text{E}_{\textbf{U}\sim P_U}[\log (1- D(G(\textbf{F}, \textbf{U}, {STTG}), \textbf{M}))]
\end{split}
\end{equation}
\noindent where %${V(G, D)}$ 
$\mathcal{L}_{G,D}$ is the binary cross-entropy loss.

\subsection{STORM-GAN Training and Testing}
% In the following, we first describe the general MAML updates and fine-tuning steps we used, and then present more specific details for STORM-GAN.

\subsubsection{MAML-based Outer Loop Updates}\label{sec:maml}
% \subsection{Cross-City Spatio-Temporal Knowledge Transfer}
As defined in Sec. \ref{sec:introduction}, our goal is to learn the shared knowledge or initialization across tasks drawn from multiple cities. 
To transfer the structural knowledge from graph and spatio-temporal knowledge from mobility data in multiple cities, we adopt the model-agnostic meta-learning (MAML) framework to learn the meta parameter $\theta_D$ and $\theta_G$, specifically in our case for all spatio-temporal tasks. The learned initialization is expected to contain common knowledge that can be fast-adapted to new tasks.
% spatio-temporal dependencies from different cities, 
% we can apply the knowledge to a unseen task from a new city with limited data. 
% $\mathcal{T}_i$
% More specifically, we first 
With MAML, we sample a batch of tasks in each step, where each task $\mathcal{T}_i$ consists of $\text{(F, M)}$ and their corresponding one-hop subgraph in STTG. 
% During task-specific estimation training, we perform the stochastic gradient descent on $\mathcal{D}_{i}^{train}$ to calculate the training loss for one task $\mathcal{T}_i$:    
The general optimization formulation is as follows.
% Our general meta-learning framework is defined as follows: 
Given a set of tasks $\left \{ \mathcal{T}_1, \mathcal{T}_2, ...  \right \}$ drawn from a task distribution $\mathit{p\left ( \mathcal{T}  \right )}$, 
% The task space ${\mathcal{T}}$ is divided into a meta-training tasks ${\mathcal{T}}_{train}$ and a meta-testing tasks ${\mathcal{T}_{test}}$.
% , and ${\mathcal{T}}_{train} \cap {\mathcal{T}_{test}}$ = \O, which means data sampled in the meta-test tasks are unseen during the training phase of meta-learning. 
where each task $\mathit{\mathcal{T}_i} \sim  p(\mathcal{T})$ consists of a training and a test set $\{\mathit{\mathcal{D}_{i}^{train}}, \mathit{\mathcal{D}_{i}^{test}}\}$,  we optimize the $G$ and $D$ with parameters $\theta_{G}$ and $\theta_{D}$ to minimize the expected empirical loss across all tasks during meta-training. The meta-update rules are given by:
\vspace{-10pt}
\begin{equation}\label{eq:maml3}
% \small
    \theta_D  = \theta_D - \beta  \nabla_{\theta_D}\mathcal{L}_{G,D}(f_{\theta'_{D}})
    %^{test}
\end{equation}
\vspace{-5pt}
\begin{equation}\label{eq:maml4}
% \small
    \theta_G  = \theta_G - \beta  \nabla_{\theta_G}\mathcal{L}_{G,D}(f_{\theta'_{G}})
    %^{test}
\end{equation}
\noindent where $\beta$ is the learning rate for meta-update, and $\theta'_G$ and $\theta'_D$ represent temporary task-specific parameters. Following the recommendation in \cite{finn2018probabilistic}, we use the first-order MAML for meta-weight update.
%A meta-testing phase is used to evaluate the meta-performance of the learned initialization. Following the same strategy in \cite{finn2017model}, 
% The tasks used for meta-testing are held out from meta-training.
% The same iteration is applied on updating meta-$\theta$ every time new batch are sampled. During the meta-testing phase, the same procedure is applied using ST-GAN model structure. The optimal $\theta^{*}$ is expected to well-learn a shared knowledge across all meta-training tasks and can be effectively adapted by unseen tasks from new cities. 

% During the fine-tuning phase, %(e.g., updating the optimal initialization $\theta^*$ for a new task from a new city), 
% we use available training samples from the new task to perform gradient descent on STORM-GAN using the meta-parameter $\theta_D$ and $\theta_G$ as the initialization for fast-adaptation.
% Finally, for fine-tuning, we use training samples from a new task (e.g., a new city) to perform gradient descent on STORM-GAN using the learned meta-parameters $\theta^*$ (Eq. \ref{eq:mamlloss}) as the initialization.

\subsubsection{STORM-GAN Inner Loop Updates}
% , and both ${G}$ and ${D}$ are trained backpropagating the loss in Eq. \ref{eq:obj} through respective models to update their parameters. 

The detailed meta-training procedure is shown in Alg. \ref{alg:training}. The training of discriminator uses the three types of combinations: ($\textbf {M}'_G$, $\textbf{F}$), ($\textbf{M}_{real}$, $\textbf{F}$) and ($\textbf{M}_{mismatch}$, $\textbf{F}$). Denote $\alpha$ as the learning rate of discriminator, $\theta'_D$ as the parameters of discriminator, the loss function and the update rule of $D$ are shown in Eq. (\ref{eq:dloss}) and Eq. (\ref{eq:dupdate}), respectively. 
% \boldsymbol
\begin{equation}\label{eq:dloss}
\small
\begin{aligned}
    f_D =& - \frac{1}{m}\sum_{i=1}^{m} \bigg(
    \log(1 - D((\textbf{M}'_G)^i, \textbf{F}^i))
    + \log(D(\textbf{M}^i_{real}, \textbf{F}^i))\\
    & + \log(1 - D(\textbf{M}^i_{mismatch}, \textbf{F}^i))
    \bigg)
\end{aligned} 
\end{equation}
% \vspace{-10pt}
\begin{equation}\label{eq:dupdate}
\small
\theta'_D = \theta'_D - \alpha\nabla f_D(\theta'_D)
\end{equation}
where $m$ is the total number of samples in a batch, and index $i$ refers to the $i^{th}$ sample. Denote $\theta'_G$ as the parameters in $G$, we have the loss function and update rule of $G$ as:
\begin{equation}\label{eq:gloss}
\small
\begin{aligned}
    f_G &= \frac{1}{m}\sum_{i=1}^{m} \bigg(
    \log(1 - D((\textbf{M}'_G)^i, \textbf{F}^i))
    \bigg)\\
    &= \frac{1}{m}\sum_{i=1}^{m} \bigg(
    \log(1 - D(G(\textbf{F}_i, \textbf{U}_i, {STTG}), \textbf{F}^i))
    \bigg)
\end{aligned}
\end{equation}
% \vspace{-5pt}
\begin{equation}\label{eq:gupdate}
\small
    \theta'_G = \theta'_G + \alpha\nabla f_G(\theta'_G)
\end{equation}

% For We sample a batch of tasks from the task distribution, we sum up the estimation loss for all tasks use $\mathcal{D}^i_{test}$, and update meta-$\theta$ once, the loss function is defined as: 
% \begin{equation}\label{eq:loss2}
%     \mathcal{L}_i^{test} = \frac{1}{m}\sum_{i=1}^{m}\left ( {M}^i_\text{real} - {M}'_G \right )^2
% \end{equation}

% Finally, we evaluate the model on meta-testing dataset ${\mathcal{T}_{test}}$ by first fine-tuning the $f_{\theta ^*}$ using $\mathcal{D}^i_{train}$, and then feed $\mathcal{D}^i_{test}$ to generate the a series of human mobility maps.  

\subsubsection{STORM-GAN Adaptation on New Tasks}
During the model adaptation phase (e.g., updating the optimal initialization for a new task from a new city), we first copy $\theta_{G}$ and $\theta_{D}$ from the meta-training phase as the initialization for fast-adaptation, and then use training samples from the new task to perform STORM-GAN for updating the meta-parameter $\theta_D$ and $\theta_G$. Finally, the updated model outputs the estimated mobility using testing samples. The tasks used for meta-testing adaptation are held out from meta-training.  

\begin{algorithm2e}
	\caption{STORM-GAN Training and Testing}\label{alg:training}
	\begin{algorithmic}[1]
		\REQUIRE \quad\\
		$\bullet$ Set of training cities $\textbf{T}_\text{train}$; set of testing cities $\textbf{T}_\text{test}$ \\
		$\bullet$ Conditions $\textbf{F}$, mobility $\textbf{M}_\text{real}$, a STTG $\mathcal{G}$\\
		$\bullet$ Inner learning rate $\alpha$; outer learning rate $\beta$; number of epochs $epoch$\\
		%$\bullet$ Number of epochs $epoch$ \\
		%$\bullet$ Graph PPG \\
	    \ENSURE $\theta_G$, $\theta_G$, estimated mobility $M'_{G}$ for $\textbf{T}_\text{test}$\\
		\COMMENT {\# Meta-learning on training cities}
		\STATE G = initG(); D = initD()
		\STATE Randomly initialize meta $\theta_G$, $\theta_G$
		\FOR{$e$ = 1 to $epoch$}
		    \STATE Sample a batch of $\mathcal{T}$ from $\textbf{T}_\text{train}$ 
		    \STATE Sample the subgragh $\textbf{H}$ of $\mathcal{T}$ from $\mathcal{G}$
		  \FOR{$\mathcal{T}_i$ in $\{\textbf{F}, \textbf{M}_\text{real}, $\textbf{H}$ \}$}
		    \STATE Sample a set of disjoint $\mathit{\mathcal{D}_{i}^{train}}$, $\mathit{\mathcal{D}_{i}^{test}}$
		    %  \COMMENT{\#discriminator}
		    \STATE Generate graph embedding $\textbf{E}$ of $\textbf{H}$\\
		    \STATE $\textbf{M}'_G$ = G($\textbf{F}$, $\textbf{E}$, $rand$($P_\textbf{U}$))
		    \STATE Update D using $\mathit{\mathcal{D}_{i}^{train}}$ by Eqs. (\ref{eq:dloss}) and (\ref{eq:dupdate})
		  %  \STATE $D_{input}$ = concat($\textbf{M}'_G$, batch.$\textbf{C}$)
		  %  D.update()
    		  %  \COMMENT{\#generator}
    % 		\STATE \{\# for generator:\}
    		\STATE Update G using $\mathit{\mathcal{D}_{i}^{train}}$ by Eqs. (\ref{eq:gloss}) and (\ref{eq:gupdate})
    		\STATE Evaluate estimation loss using $\mathit{\mathcal{D}_{i}^{test}}$ by Eq. (\ref{eq:obj})
    % 		\STATE Evaluate estimation loss by $\textbf{D}_{test}$
		  \ENDFOR
		  \STATE Update $\theta_D$ and $\theta_G$ by Eq. (\ref{eq:maml3}) and Eq. (\ref{eq:maml4})
		  %\STATE updateLearningRate($r$)
		\ENDFOR \\
		\STATE Return $\theta_G$, $\theta_D$ \\
		%\vspace{3pt}
		\COMMENT {\# Fast-adaptation on testing cities}
		\STATE Sample batch of testing tasks $\mathcal{T}$ from $\textbf{T}_\text{test}$ 
	  \FOR{$\mathcal{T}_i$ in $\{\textbf{F}, \textbf{M}_\text{real}, $\textbf{H}$ \}$}
		\STATE Sample a disjoint $\mathit{\mathcal{D}_{i}^{train}}$, $\mathit{\mathcal{D}_{i}^{test}}$ from $\textbf{T}_\text{test}$\\
		\STATE Generate graph embedding $\textbf{E}$ of $\textbf{H}$\\
		\STATE Copy $\theta_G$, $\theta_D$
		\STATE Evaluate performance by Eq. (\ref{eq:obj}) using $\mathit{\mathcal{D}_{i}^{train}}$
		\STATE Update G through Eqs. (\ref{eq:gloss}) and (\ref{eq:gupdate})
		\STATE Estimate $\textbf{M}'_G$ using updated G and $\mathit{\mathcal{D}_{i}^{test}}$
	  \ENDFOR
		
	% \vspace{-5pt}
	\end{algorithmic}
\end{algorithm2e}

\section{Evaluation}\label{sec:evalution}
% Through the experiments we aim to answer the following questions:
% \begin{itemize}[leftmargin=*]
%     \item Whether STORM-GAN can outperform baseline methods in terms of solution quality?
%     \item How does the proposed spatio-temporal network impact the solution quality compared to non-spatial-temporal models? 
%     \item What is the effect of the meta-learning framework on model performance?
%     \item Can the proposed STTG embedding contribute to the model performance?
% \end{itemize}

\subsection{Dataset Description}
%The first column in fig. \ref{fig:validation} shows the data categories we gathered from multiple sources based on definitions in Sec. \ref{sec:concept} (i.e., contextual, epidemic and policy conditions; human mobility responses).
% \noindent{\bf Data Sources.} We collect four mobility related datasets (i.e., contextual, epidemic, policy, and human mobility responses) from heterogeneous sources (i.e., Census\cite{acs}, US CDC \cite{cdc}, and SafeGraph mobility patterns \cite{safegraph}). SafeGraph provides free access to data for academic purpose with upon request and all the other data are publicly available. Policy 
% The dataset spans over six cities in different states located from the west coast, midwest to the east coast (i.e., Boston, Chicago, Houston, Iowa City, Los Angeles, and New York City). 
% % to evaluate the performance of our proposed method. 
% The list of cities also covers regions from large metropolitan areas to less populous places. The candidate methods are trained on five cities (meta-training) with one left out as the new city for meta-testing. %Specifically, we selected Houston (large metropolitan area) and Iowa City (small urban area) as the two test cities in two separate experiments. 
% Adaptation on test cities is performed with data samples from the most recent two weeks (out of 35 weeks in total). 

\noindent{\bf Data Sources.} We elaborate four types of data as described in Def. \ref{def:conditions} \& Def. \ref{def:reponses} (pandemic, contextual, policy, and mobility). They are collected from Centers for Disease Control and Prevention \cite{cdc}, Census Bureau \cite{acs}, the date of disease prevention policies were collected from the corresponding city government website news, and SafeGraph \cite{safegraph}, respectively. SafeGraph provides free access to data for academic purposes with upon request and all the other data are publicly available. 

\begin{figure}
  \centering
  \includegraphics[scale=0.555]{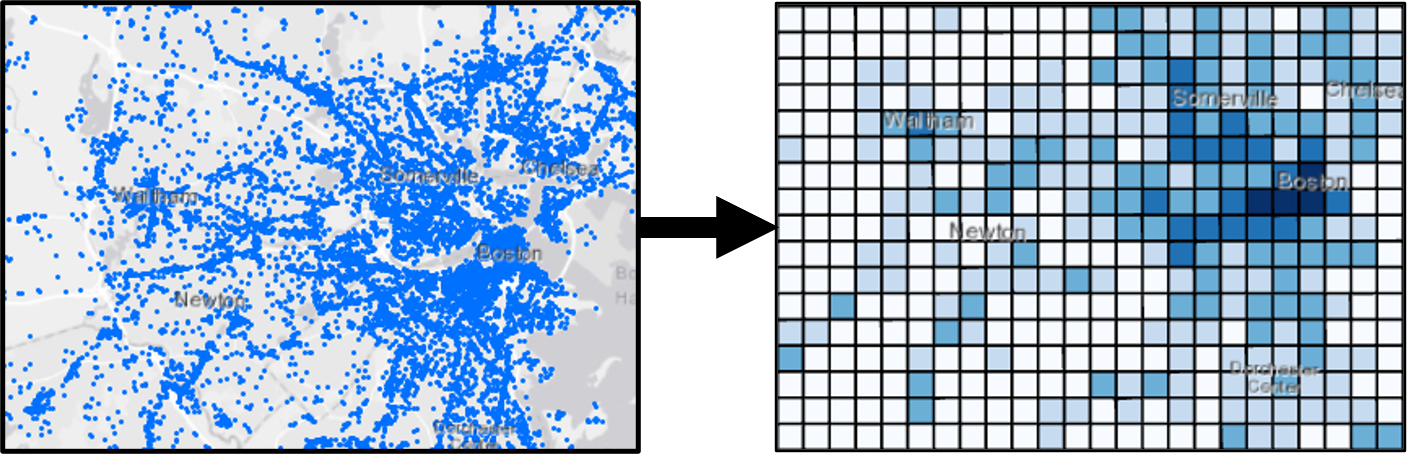}
  % \vspace{-2pt}
  \caption{POI visit counts data pre-processing.\vspace{-10pt}}
  \label{fig:description}
\end{figure}

\noindent{\bf Data Granularity.} The original POI dataset from SafeGraph is obtained by collecting the location from cell phone records with latitude and longitude information. Then, the location information is used to determine the visits to POIs \cite{safegraph}. The POI visit counts data is in point data format. Fig. \ref{fig:description} illustrates the discretization of one city, where we sum the total POI visit counts that fall into each grid cell, and use this aggregated visit counts value to represents the human mobility responses of each grid cell. 

\noindent{\bf Data Pre-processing.} To construct the list of conditions for our input, for each grid cell, we preprocess data collected from different sources with different geographic units. We first adopt a commonly used space-partitioning method to segment each spatial domain into grid cells of size of 1km $\times$ 1km, and segment all mobility related conditions using the same grid cells. Then, each spatial region (or unit spatial window) we used to create a data sample is a $10\times 10$ spatial window on the grid. For each grid cell, the value of human mobility response is the total number of POI visit counts in a day. Note that some conditions are re-scaled during this process. For example, population and median household income data are collected at the census tract level, and we linearly re-scaled the data using the corresponding area ratios between the area of the original census tract polygon and the proposed $10\times 10$ grid cells. Similarly, COVID-19 statistics and policy data are collected at the county level. We assign each grid cell with the corresponding data on which county it belongs.

\noindent{\bf Training Data Description.} 
We collect the mobility related datasets from six cities. The dataset spans over six cities in different states located from the west coast, midwest to the east coast (i.e., Boston, Chicago, Houston, Iowa City, Los Angeles, and NYC). 
% to evaluate the performance of our proposed method. 
The list of cities also covers regions from large metropolitan areas to less populous places. Detailed statistics of these datasets (e.g., number of POIs, number of cells covered for each city) are listed in Table \ref{tab:data2} and Table \ref{tab:poi}. The duration of data is from 02/24/2020 to 10/25/2020, covering 35 weeks in total. As discussed in Methodology section, the data is segmented into a spatio-temporal distribution of tasks, where each task contains one single city for five consecutive weeks (no mutual overlaps among tasks).
The candidate methods are trained on five cities (meta-training) with one left out as the new city for meta-testing. %Specifically, we selected Houston (large metropolitan area) and Iowa City (small urban area) as the two test cities in two separate experiments. 
Adaptation on test cities is performed with data samples from the most recent two weeks (out of 35 weeks in total). 
% As the duration of data is from 02/24/2020 to 10/25/2020, covering 35 weeks in total, we divide each city into 7 tasks along temporal dimension, and each task maintains the full spatial domain of corresponding city. 
Overall, we have $\textbf{35}$ spatio-temporal estimation tasks in total. For methods with meta-learning, 80\% of data in each task is used for meta-training, and the rest for testing (Def. \ref{def:sttask}).
\vspace{-5pt}
\begin{table}[h]
\centering
\caption{Detailed data statistics.}
\vspace{-5pt}
\label{tab:data2}
\resizebox{\columnwidth}{!}{%
\begin{tabular}{|c|c|c|c|}
% |p{1.5cm}|p{2cm}|p{1cm}|p{1cm}|
\hline
City      & Time Span (M/D/Y) & POIs & Size          \\ \hline
Boston & \multirow{6}{*}{\begin{tabular}[c]{@{}c@{}}02/24/2020\\ -10/25/2020\\ \\ 35 Weeks\end{tabular}} & 26054 & 37 $\times$ 48 \\ \cline{1-1} \cline{3-4} 
NYC       &          & 133520  & 58 $\times$ 72 \\ \cline{1-1} \cline{3-4} 
LA        &          & 86721 & 52 $\times$ 64   \\ \cline{1-1} \cline{3-4} 
Chicago   &          &  47356  &   50 $\times$ 40 \\ \cline{1-1} \cline{3-4} 
Houston   &          &  37315  & 50 $\times$ 60 \\ \cline{1-1} \cline{3-4} 
Iowa City &          &   1401  & 20 $\times$ 32 \\ \hline
\end{tabular}}%
\end{table}
\vspace{-5pt}
\begin{table}[h]
\centering
\caption{Average number of POI visit counts per grid cell}
\vspace{-5pt}
\label{tab:poi}
\resizebox{\columnwidth}{!}{%
\begin{tabular}{|c|c|c|c|c|c|c|}
\hline
City      & Boston & NYC & LA & Chicago & Houston & Iowa City       \\ \hline
POI Counts       &     28     & 64  & 52 & 30 & 24 & 6 \\  \hline 
\end{tabular}}%
\end{table}

\subsection{Evaluation Metrics}\label{sec:metrix}
We evaluate the performance of STORM-GAN by using the following measures: mean absolute error (MAE)
%: $MAE = \frac{1}{n} \sum_{i=0}^{n} \left | M'_G - {M}_{real}\right |$ 
and rooted mean square error (RMSE).
%: $RMSE = \sqrt{\frac{1}{n}\sum_{i=1}^{n}\left ( M'_G - {M}_{real} \right )^2}$.
%where $M_G$ is the real mobility response and $\hat{M_G}$ is the generated mobility response values by candidate approach.  %\begin{equation}\label{eq:mae}
\begin{equation}\label{eq:mae}
\small
\begin{split}
    MAE = \frac{1}{n} \sum_{i=0}^{n} \left | M_G - \hat{M_G}\right |
\end{split}
\end{equation}

\vspace{-5pt}

\begin{equation}\label{eq:rmse}
\small
\begin{split}
    RMSE = \sqrt{\frac{1}{n}\sum_{i=1}^{n}\left ( M_G - \hat{M_G} \right )^2}
\end{split}
\end{equation}
where $M_G$ is the real mobility response and $\hat{M_G}$ is the generated mobility response values by candidate approach. Since model generates the spatial unit windows multiple times for each grid cell during estimation, the outputs of generator are averaged before comparing with the ground truth. 

To evaluate the model performance of learning the data distribution, we calculate the KL divergence to indicate the similarity between the learned human mobility responses distribution $\hat{\textbf P}$ and real human mobility responses distribution $\textbf {P}$ on different bin sizes. The KL divergence is defined as follow:
% using Houston as testing city 
\begin{equation}\label{eq:mae}
\begin{split}
% \small
    \mathbf{D_K{}_L}(P||\hat{P}) = \sum_{i=1}^{N}P(M'_G)log(\frac{P(M'_G)}{\hat{P}({M}_{real})})
\end{split}
\end{equation}
% we obtain the two distributions by calculating $\textbf {M}'_G$ and $\textbf{M}_\text{real}$ histograms, and then applying different bin sizes to evaluate the similarity between ground truth distribution and learned distribution.

% \subsection{Experiment Setting}\label{sec:setting}
% % As we formulate the estimation problem as generating a 7-day human mobility responses maps conditioned on related features, 

% %(e.g., updating the optimal initialization $\theta^*$ for a new task from a new city 

% %add constraint layer information here. 

% %implement constraint layer
\hspace{5pt}
\subsection{Baseline Methods}\label{sec:base}
% We design two categories of baselines: without meta-learning framework and with meta-learning framework. During training, for model without meta-learning framework, we feed all data from meta-training tasks to train the model and use meta-testing data to evaluate without updating meta parameter. For models with meta-learning framework, we learn the meta-parameter from meta-training tasks and adapt to unseen tasks to improve the estimation performance. We compare our proposed framework with the following baseline models:
We compare our proposed method with the following baseline methods, and fine-tune each method using Houston and Iowa City as testing cities respectively.
%, exclude historical average and spatial smoothing method: 
% \subsubsection{Baseline without meta-learning framework}
\begin{itemize}[leftmargin=*]
  \item $\textbf {HA}$: Historical Average. The average of human mobility responses calculated using observed values from the same location in the past two weeks (same weekday). 
%   In this paper, we calculate the historical average using mobility from unseen task. 
  \item \textbf{Spatial smoothing with neighborhood regions} \cite{getis2008history}. This method uses the mobility response values in a local $3\times 3$ window to compute a mean as the estimated value. The values for smoothing are from the same weekday in the most recent week.
  %The values for smoothing are from the same weekday in the most recent week.
%   Note we use training set in meta-testing task to estimate for this method.
  \item \textbf{Ridge} \cite{hoerl2020ridge}. We use ridge regression with the same input features and mobility responses.
  \item \textbf{cGAN} \cite{zhang2019trafficgan}. A conditional GAN where the generator and discriminator use three fully-connected layers (no layer structure to learn spatial or temporal patterns).
%   A standard conditional Generative Adversarial Network (cGAN). 
%   , and the output of the generator is activated by Tanh, and the output of discriminator is fed to Sigmoid function.
%   \item $\textbf {COVID-GAN} \cite{10.1145/3397536.3422261}$: COVID-GAN, a spatio-temporal Conditional Generative Adversarial Network for estimating human mobility during COVID-19.
% \end{itemize}
% \subsubsection{Baseline w. meta-learning framework}
% \begin{itemize}
  \item \textbf{COVID-GAN} \cite{10.1145/3397536.3422261}: COVID-GAN has the same structure as the above cGAN, and it adds a correction layer, which is used to add constraints based on policy to refine the results. %\cite{10.1145/3397536.3422261}.
%   We apply a meta-learning framework on COVID-GAN's outer structure.  
  \item \textbf{MAML-DAWSON} \cite{liang2020dawson}: An optimization-based meta-learning approach using MAML. As DAWSON originally works on music generation tasks, we modify its inner structure with a regression-focused conditional GAN.
%   to serve as estimation purpose. 
  \item \textbf{MetaST} \cite{yao2019learning}: MetaST fuses CNN, LSTM and attention mechanism to predict urban traffic volume through MAML framework. 
\end{itemize}

\subsection{STTG Construction Examples}\label{sec:sttg}
In this section, we provide two different STTG construction scenarios to evaluate the effectiveness of graph embedding in human mobility estimation. 

\emph{Scenario 1 (S1).}
%Among metropolitan areas, the human mobility responses auto-correlation is heterogeneous, varying from the geographic location to the different urban land-use patterns. For example, 
We assume that cities of similar sizes, socio-economic environment, and land-use design may share similar human mobility pattern that could help the estimation of new cities. To build the graph $\mathcal{G}_{s1}\left ( \mathcal{V},\mathcal{E}  \right )$, we enumerate major metropolitan cities from every region in U.S., and define each city as a node $\nu_i$. Next, we extract human mobility maps for all the cities from a same date, and calculate the pairwise distribution similarity score between cities using KL-divergence. The KL-divergence indicates the strength of human mobility correlation. Each edge is added if the correlation is $\leqq$ 0.5, and is weighted by the correlation. $\mathcal{G}_{s1}$ contains 55 nodes and 682 edges. Node attribute store the outbreak stage of COVID-19. Each stage value is in $\{1,2,3\}$, where a smaller value means earlier in COVID-19 outbreak. The stage value is assigned based on the month when an exponential growth is first appeared. 
%node attribute which is the correlation vector. %\xz{I didn't get this. Node attribute is the matrix???}  

\emph{Scenario 2 (S2).} 
Intuitively, urban environment increases the chance of infection as people move around and interact with others and the environment. As a hub for migration and travel, urban areas may quickly spread infections to nearby places through short-distance travel, and to major cities through connection flights. 
% We design a graph representing the infection transmission in United States to capture the spatial-temporal dependency among different cities.
% % to facilitate the learning in related tasks. 
% % nearby regions through short-distance travel, and propagate

To construct ${S2}$, similar to ${S1}$, we enumerate major metropolitan areas in U.S, and define a graph $\mathcal{G}_{s2}\left ( \mathcal{V},\mathcal{E}  \right )$ to represent the relationships of these cities. 
% The nodes in the graph $\nu \in \mathcal{V}$ are cities that are connected to those in the input data.
% In the graph, each node $\nu \in \mathcal{V}$ denotes a city 
% which should satisfy the following requirements: (1) it contains an international or large airport; (2) it is related to the input conditions ${F}$ so that the utilization of ST-GAN does not depend on constraint. 
We divide the nodes $\mathcal{V}$ into two categories: the hub nodes $\mathcal{V}^h$ are major cities with more than 100 airlines; the second-tier nodes $\mathcal{V}^s$ are cities with more than 35 but less than 100 airlines. Moreover, each directed edge $\nu_i \to \nu_j \in \mathcal{E}$ is added if its two nodes are: (1) both major cities that have direct flights; or (2)  within a spatial proximity threshold (500 km in this paper). Our graph contains 69 nodes and 776 edges. 
% Thus, the edge rules we use here for PPG construction assume that the disease transmit through airlines or neighborhood cities will affect each other due to more active human movement. 

\begin{figure}
  \centering
  \includegraphics[scale=0.38]{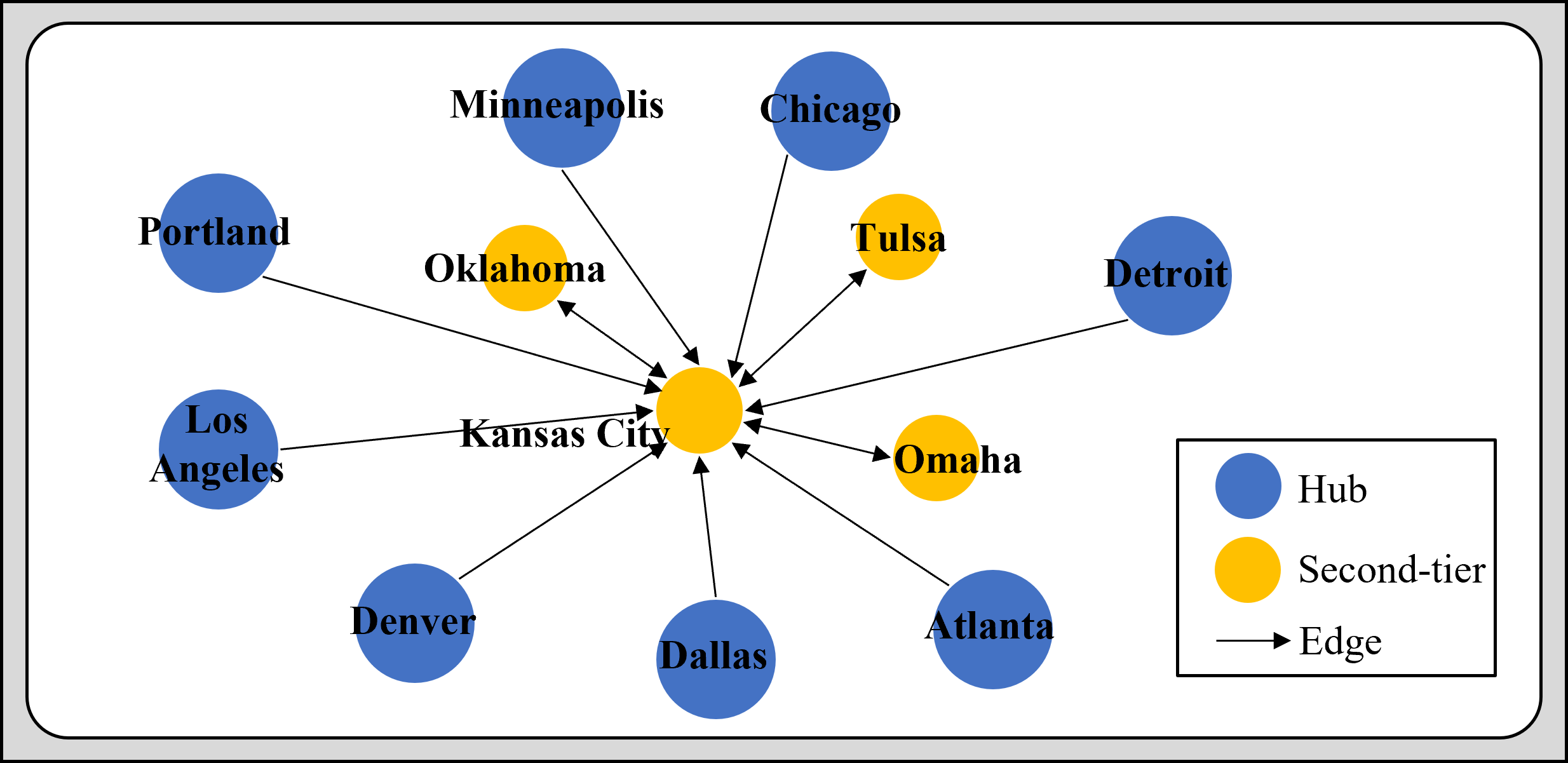}
  \captionsetup{justification=centering}  
  \caption{Subgraph of a node in STTG \emph{S2}.
  }
  \label{fig:graph}
  \vspace{-10pt}
\end{figure}

The graph is then weighted by spatio-temporal attributes associated with nodes and edges. Edge attributes contain the number of directed flights between the cities and their geographic distance.
% ; for edges without direct flights, the attribute is dominated by geographic distance. 
Node attributes store the sum of flights from connected edges as well as the outbreak stage of COVID-19 which is the same as scenario 1. %Each stage value is in $\{1,2,3\}$, where a smaller value means earlier in COVID-19 outbreak. The stage value is assigned based on the month when an exponential growth is first appeared. 
Fig. \ref{fig:graph} shows an illustration example of our ${S2}$ which is a 1-hop subgraph for Kansas City, a second-tier city by the above-mentioned classification. 
% For all the international airports with more than 100 airlines, we extract 
Major cities that have direct flights to Kansas City (e.g., Denver, Atlanta, Minneapolis) and second-tier cities (e.g., Oklahoma, Omaha) within the spatial proximity threshold are shown on the subgraph.

We use both of the two STTG definitions with our STORM-GAN (namely, STORM-GAN($S1$) and STORM-GAN($S2$)) to evaluate its performance in the next subsection.

\subsection{Estimation Quality Evaluation}\label{sec:evaluation results}
The benefit of meta-learning is that the model can quickly update the model parameters, and generate good results on a new task by seeing a small fraction of new data. During the adaptation phase, we evaluate the performance of the candidate methods on the two test cities (i.e., Houston and Iowa City) using their last two weeks (Monday to Sunday) of data.
The length of a time period we use is 7-day since human mobility pattern is influenced by strong weekly periodicity. For each testing city and week, we use 2 consecutive weeks of data ahead of the week for adaptation, and then use the parameters to generate the next 7-day (one week) human mobility responses.
%for a longer testing period. %(unless specified specifically in Sec. \ref{sec:base}). 
%Furthermore, for all generated mobility maps, we apply the correction filtering proposed in \cite{10.1145/3397536.3422261}, which helps remove all mobility responses at cells with no valid POIs.% (i.e., no POI or no POIs that are open according to the policy feature).
% after generating the estimated mobility maps to mitigate spurious results during training.  
% \usepackage{graphicx}

% \usepackage{graphicx}

\begin{table*}[]
%\begin{tabular}{|p{1.5cm}|p{5.5cm}|}
\small
\centering
\caption{Human mobility responses estimation by candidate methods for Houston 
% (7-day window)
% ($^*$difference to best$\leq0.1$)
\vspace{-5pt}
}
\label{tab:result}
% \resizebox{\textwidth}{!}{%
\resizebox{\textwidth}{!}{%
\begin{tabular}{c|ccccccc|ccccccc} \hline
      & \multicolumn{7}{c|}{RMSE}                             & \multicolumn{7}{c}{MAE}                              \\
               \hline%\cline{2-8} \cline{9-16}
             Model  & Mon & Tues  & Wed  & Thu & Fri & Sat & Sun & Mon & Tues & Wed & Thu & Fri & Sat & Sun \\ \hline
HA             & 194.8   & 193.2   & 193.5   & 196.3    & 193.6   & 194.5   & 195.1   & 81.2    & 80.9    & 80.4    & 80.8    & 79.6    & 81.3    & 81.2    \\
Smoothing      & 150.9   & 168.1   & 169.2   & 177.1   & 187.3   & 202.4   & 162.1   & 82.3    & 90.2    & 90.4    & 94.6    & 100.1   & 99.8    & 104.2   \\
cGAN           & 278.2   & 283.4   & 284.6   & 279.5   & 286.3   & 286.1   & 286.2   & 118.7   & 122.4   & 125.3   & 120.6   & 128.3   & 130.6   & 129.3   \\
Ridge          & 189.4   & 192.1   & 182.3   & 181.5   & 187.3   & 188.6   & 195.7   & 95.8    & 99.5    & 101.9   & 95.3    & 99.4    & 98.1    & 95.2    \\
% COVID-GAN      & 202   & 218   & 214   & 215   & 219   & 205   & 208   & 88    & 92    & 92    & 91    & 86    & 91    & 90    \\
COVID-GAN & 171.5   & 175.6   & 174.3   & 172.2   & 176.4   & 171.8   & 170.6   & 75.1    & 80.1    & 73.2    & 77.1    & 80.5    & 81.7    & 82.8    \\
MAML-DAWSON  & 169.5   & 169.7   & 166.1   & 168.9   & 164.6   & 166.7   & 168.1   & 68.3    & 67.4    & 70.3    & 69.4    & 68.2    & 75.5    & 78.2    \\
MetaST & 170.2  & 171.4 & 173.8 & 170.6 &  169.5 & 170.3 & 169.4 &  72.8 &  76.1 & 71.5 & 74.2 & 72.9 &  80.2 & 81.3  \\
STROM-GAN (${S1}$)  & 151.2 & 150.5 & 149.3 & 162.9 & 163.9 & 164.2 & 167.2& 66.7 & 66.4 & 64.7 & 63.1 & 70.4 & 71.8 & 72.2 \\
STORM-GAN (${S2}$)    & \textbf{145.1}   & \textbf{142.6}   & \textbf{141.9}   & \textbf{141.6}   & \textbf{152.5}   & \textbf{156.7}   & \textbf{160.2}   & \textbf{61.7}    & \textbf{60.4}    & \textbf{59.3}    & \textbf{53.8}    & \textbf{58.4}    & \textbf{64.2}    & \textbf{67.2}   \\ \hline
\end{tabular}}%
% }
% \vspace{-10pt}
\end{table*}

\begin{table*}[]
\small
\centering
\caption{Human mobility responses estimation by candidate methods for Iowa City %(7-day window)
% ($^*$difference to best$\leq0.1$)
\vspace{-5pt}
}
\label{tab:iowa}
% \resizebox{\textwidth}{!}{%
\begin{tabular}{c|ccccccc|ccccccc} \hline
               & \multicolumn{7}{c|}{RMSE}                              & \multicolumn{7}{c}{MAE}                               \\\hline
Model & Mon & Tues  & Wed  & Thu & Fri & Sat & Sun & Mon & Tues & Wed & Thu & Fri & Sat & Sun\\
% Day 1 & Day 2 & Day 3 & Day 4 & Day 5 & Day 6 & Day 7 & Day 1 & Day 2 & Day 3 & Day 4 & Day 5 & Day 6 & Day 7 \\ 
\hline
HA             & 22.2    & 23.1    & 21.3    & 24.2    & 22.3    & 25.2    & 26.5    & 13.4    & 13.1    & 15.2    & 12.6    & 14.2    & 16.1    & 13.2    \\
Smoothing      & 18.4    & 16.3    & 18.6    & 21.7    & 18.3    & 19.2    & 19.5    & 11.2    & 10.3    & 10.6    & 9.1     & 11.6    & 9.2     & 11.6    \\
cGAN           & 34.3    & 33.7    & 34.8    & 36.5    & 32.8    & 31.2    & 34.4    & 21.1    & 19.5    & 21.3    & 18.7    & 17.8    & 19.9    & 20.3    \\
Ridge          & 20.6    & 22.3    & 21.1    & 20.5    & 19.8    & 19.2    & 20.1    & 12.4    & 11.6    & 13.3    & 13.6    & 12.2    & 13.5    & 12.2    \\
% COVID-GAN      & 18    & 17    & 16    & 18    & 18    & 19    & 20    & 15    & 18    & 14    & 13    & 13    & 15    & 15    \\
COVID-GAN & 17.1    & 17.3    & 16.6    & 16.3    & 15.2    & 14.3    & 15.6    & 11.3    & 10.7    & 13.1    & 12.3    & 14.7    & 13.8    & 13.3    \\
MAML-DAWSON         & 16.5    & 17.6    & 17.7    & 15.2    & 15.3    & 13.6    & 13.4    & 10.3    & 11.6    & 10.1    & 9.5     & 8.1     & 10.9    & 9.2     \\
MetaST & 17.1  & 17.4 & 17.8 & 17.6 &  16.5 & 16.3 & 16.1 & 10.8 & 10.6 & 10.4 & 10.5 & 9.9 & 10.1 & 9.7 \\
STROM-GAN (${S1}$) & 15.8 & 16.2 & 16.1 & 15.8 & 15.9 &15.4 & 14.3 & 8.9 & 9.6& 9.9 & 10.2 & 9.1 & 9.3 & 9.3  \\
STORM-GAN (${S2}$) & \textbf{14.1}    & \textbf{15.6}    & \textbf{14.9}    & \textbf{14.4}    & \textbf{14.2}    & \textbf{13.6}    & \textbf{13.3}    & \textbf{8.2}     & \textbf{7.4}     & \textbf{9.1}     & \textbf{8.3}    & \textbf{7.8}     & \textbf{9.1}     & \textbf{8.5}   \\ \hline
\end{tabular}%
% }
\vspace{-5pt}
\end{table*}

% \par \noindent 
\textbf{Performance comparison of proposed STORM-GAN and other candidate methods.} 
% We aim to answer the first question summarized at the beginning of Sec. \ref{sec:evalution}. 
Tables \ref{tab:result} and \ref{tab:iowa} show the results of the candidate methods obtained using Houston and Iowa City as the testing city, respectively. We apply $S1$ and $S2$ graph construction scenarios on STROM-GAN, respectively. The evaluation results show that both STORM-GAN scenarios overall achieve the lowest RMSE and MAE for each day in the week, with major improvements from 91.7$\%$ to 17.6$\%$.  
%for many (e.g., $>$10\%).
% The overall performance is greatly improved by comparing the RMSE and MAE with all the baseline methods. 
% We observe that the historical average and spatial smoothing method achieve reasonably good result
% perform well but not as good as the proposed model. 
% These observation reveals that the historical average and smoothing methods can provide reasonable approximation for mobility estimation due to the temporal periodic pattern, and stable urban spatial pattern. 
It is interesting to observe that historical average and spatial smoothing methods perform better than the basic cGAN, which to some degree shows the spatio-temporal auto-correlation effects.
However, these methods can mainly estimate a rough base but are limited in capturing complex spatio-temporal relationships between features and mobility responses. 
%Ridge Regression, on the other hand, considers the linear mapping from external features to mobility responses, but does not effectively leverage spatial or temporal relationships.
% achieves a better RMSE compared with HA and smoothing.  
Comparing to COVID-GAN and MAML-DAWSON, our model outperforms
% Our model performs 
COVID-GAN by 20.5\% (RMSE) and 23.5\% (MAE) on average, and MAML-DAWSON by  
% with the incorporation of CNN and LSTM substructures, PPG . 
% In addition, comparing with DAWSON, our approach obtains a 
15.1\% (RMSE) and 23.4\% (MSE). The results show that the design of spatio-temporal architecture (i.e., CNN and LSTM substructures and the STTG graph) and meta-learning adaptation can significantly improve the solution quality. Furthermore, our model achieves 14.7\%  (RMSE) and 15.2\% (MSE) better than MetaST, demonstrating that task-based graph embedding can contribute to model performance by learning the inter-task similarities. We evaluate the model performance on less populous areas using Iowa City as testing city, the POI numbers and city size of Iowa City are significantly smaller than large metropolitan areas according to Tables \ref{tab:data2} and \ref{tab:poi}. As Table \ref{tab:iowa} shows, the improvements are relatively smaller due to smaller number of POI visit counts in less populated city. However, both STORM-GAN scenarios still achieve the lowest errors in all of the testing days.

\begin{figure}
  \centering
  \includegraphics[scale=0.38]{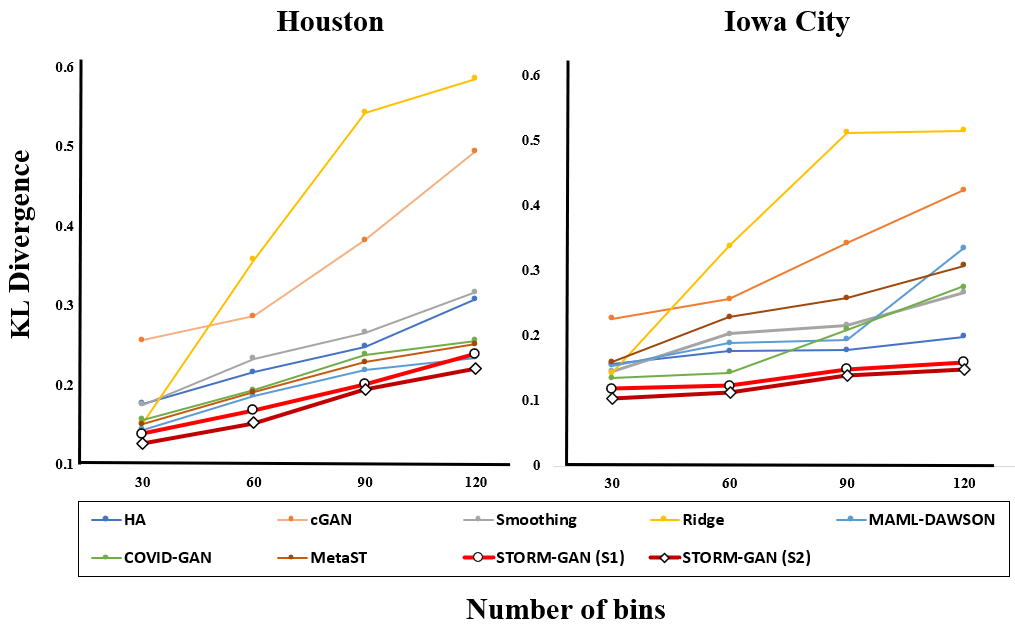}
  \caption{Kullback-Leibler divergence.
  \vspace{-10pt}
  }
  \label{fig:KL}
\end{figure}

%The distribution similarity evaluation between between real distributions $\textbf {P}$ and learned distribution $\hat{\textbf P}$ generated by candidate approaches, 
We calculate the KL-divergence using results from Houston and Iowa City (Fig. \ref{fig:KL}). The X-axis represents the number of equal-size bins used to discretize the value needed for the computation, and the Y-axis shows the KL divergence values. A lower KL divergence value means the result better matches the real distribution. As shown in Fig. \ref{fig:KL}, STORM-GAN achieves lowest KL-divergence compare to the baseline methods consistently for all numbers of bins.

\textbf{Impact of STTG choice:} Our results show that both of the two STTG constructed can significantly improve the performance of STORM-GAN in Houston and Iowa City. This proves that the spatio-temporal task-graph embedding design is effective and robust, rather than tailored for a specific STTG definition. Between the two choices, $S2$ generally achieves better performance as it uses more information that are directly related to the spreading of COVID-19. 

\begin{figure}
  \centering
  \includegraphics[scale=0.495]{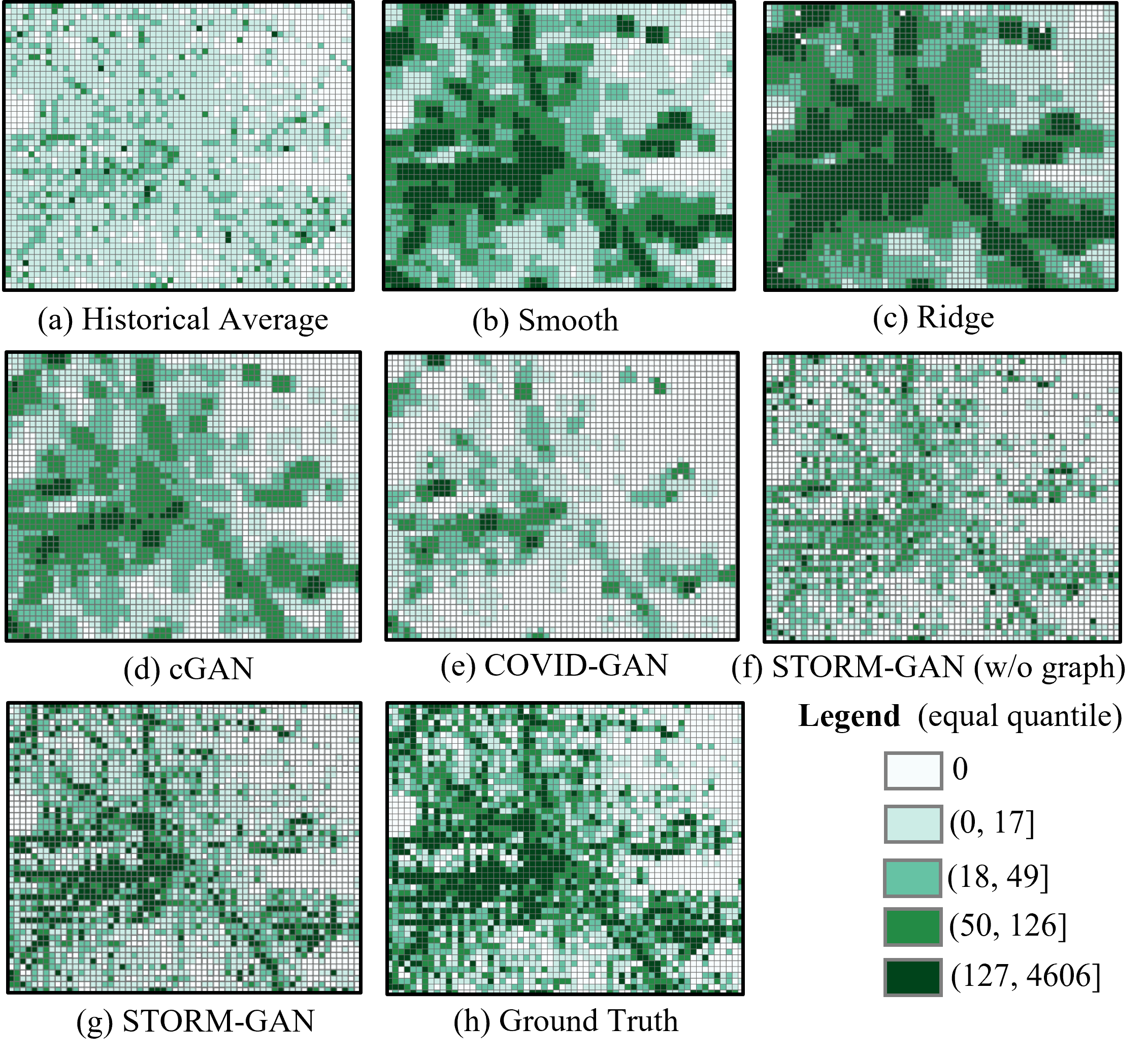}
\vspace{-5pt}
  \caption{Mobility estimation results of the Houston study area. 
\vspace{-20pt}
  }
  \label{fig:full1}
\end{figure}

\begin{figure}
  \centering
  \includegraphics[scale=0.435]{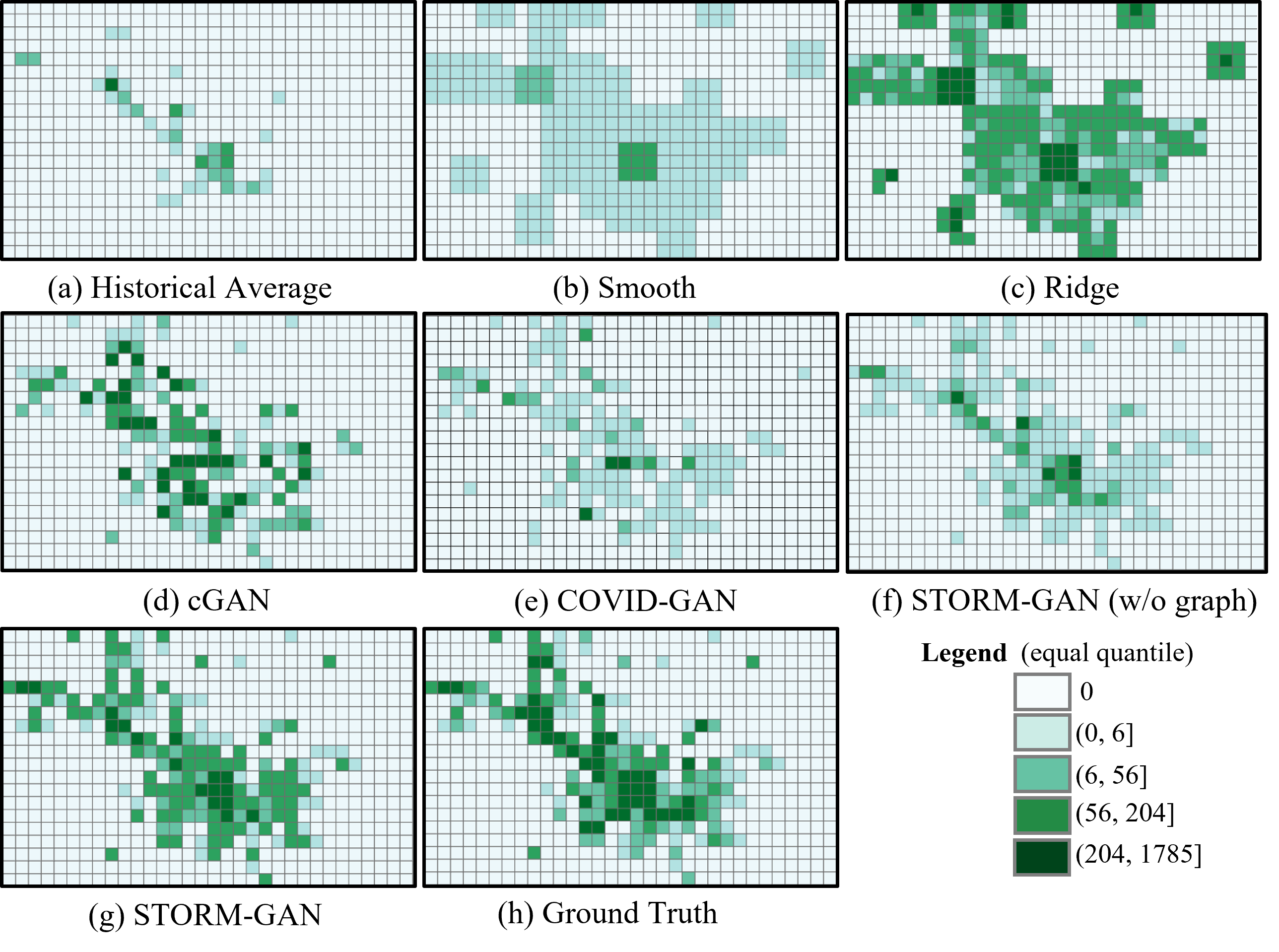}
\vspace{-5pt}
  \caption{Mobility estimation results of the Iowa City study area. 
\vspace{-20pt}
  }
  \label{fig:full2}
\end{figure}

% \par \noindent 
\textbf{Ablation Study.} We study the effect of different components proposed in our method using Houston as the testing city on one day (Monday).
% , and illustrate the results on a day:
\begin{itemize}[leftmargin=*]
  \item \textbf {Base}: Baseline conditional GAN.
  \item \textbf {Base + Spatial (S)}: Equivalent to COVID-GAN, which has a correction layer to add policy constraints, but purely a spatial model.
%   to refine the results.
%   Fine-tuned with samples from the new city. This architecture does not have any structure to learn spatial or temporal patterns. 
  \item \textbf{Base + Spatio-Temporal (ST) + Meta}: Proposed STORM-GAN with spatio-temporal meta-learning, but without the STTG graph.
%   pandemic propagation graph (PPG).
  \item \textbf{Base + ST + Meta + Graph(S2)}: Complete STORM-GAN.  \end{itemize}

\begin{table}[]
\small
\vspace{-5pt}
\centering
\caption{Comparison among STORM-GAN variations} 
\vspace{-5pt}
% \vspace{-5pt}
\label{tab:result2}
%\resizebox{\textwidth}{!}{%
\begin{tabular}{llll}
\hline
Method                         & RMSE & MAE \\ \hline
Base       & 202.2  & 80.6 \\ 
Base + S                           & 171.5  & 75.1 \\ 
Base + ST + Meta         & 149.8  & 67.6 \\ 
Base + ST + Meta + Graph (${S2}$) & 145.1  & 61.7 \\ \hline
\end{tabular}%
%}
% \vspace{-10pt}
\end{table}
% \begin{table}[]
% \small
% \centering
% \caption{Comparison among STORM-GAN variations \vspace{-5pt}}
% \label{tab:result2}
% %\resizebox{\textwidth}{!}{%
% \begin{tabular}{llll}
% \hline
% Method                         & RMSE & MAE \\ \hline
% Baseline (no fine-tuning)       & 202.2  & 88.6 \\ 
% COVID-GAN (fine-tuned)                           & 171.5  & 75.1 \\ 
% Spatio-Temporal + Meta         & 149.8  & 67.6 \\ 
% Spatio-Temporal + Meta + Graph & 145.1  & 61.7 \\ \hline
% \end{tabular}%
% %}
% \vspace{-10pt}
% \end{table}
Table \ref{tab:result2} shows the estimation performance of STORM-GAN and its variants. 
%From cGAN to STORM-GAN, the performance improve consistently, indicating that every component contributes to the solution quality. %First, we can see that STORM-GAN achieves a significant lower RMSE with reduction of 26.4\% and a lower MAE with reduction of 14\% compare to baseline method. The result demonstrates that the effectiveness of using meta-learning framework to learn knowledge across tasks. , 
First, the base + spatial (S) achieves a lower RMSE and MAE (a reduction of 15.3\% and 15.2\%, respectively) compared to cGAN, showing the effectiveness of the correction layer from COVID-GAN. Next, we can see that
the addition of spatio-temporal meta-learning further reduces RMSE and MAE by 12.7\% and 10\%, respectively.
% cGAN + correction + ST + Meta outperforms cGAN + correction with reduction of 12.7\% and 10\% in RMSE and MAE by adding the spatio-temporal and meta-learning structure. 
This result demonstrates that CNN, LSTM and meta-learning can better capture the complex spatio-temporal relationships across multiple cities. Finally, the complete STROM-GAN achieves the lowest RMSE and MAE with the sptaio-temporal task-based graph.

% \par \noindent 
\textbf{Visualization.} We compare the solution quality of seven candidate approaches through map visualization. Fig. \ref{fig:full1} and Fig. \ref{fig:full2} (a) to (f) show the results of baseline methods, and (g), (h) show the STORM-GAN (${S2}$) and ground truth. The results show the full Houston and Iowa City study areas for a day in the data. Here STORM-GAN generates fine-scale mobility values that are closer to the ground truth. As we can see, the mobility pattern generated by the STORM-GAN can capture the spatial pattern of human mobility responses better than other baselines. The reason may be that similar functionality zones at different cities may have similar mobility patterns. The meta-learning framework successfully learns this shared knowledge from training tasks. Moreover, the utilization of CNN and LSTM helps capture the spatio-temporal correlation from region to region.   

%COVID-19 related policy has a discouraging effect on people’s will to get out. Forcing the model to learn based on the conditional historical average value controlled by policy leads to stronger suppressing force on the value and an increase in the estimation. Note standard cGAN generate mobility response values in the sea region locates on the middle of the right side. This supports the usage of the domain-knowledge constraint correction mechanism.

\section{Other Related Work}
% \par \noindent
% \textbf{COVID-19.} There have been many studies \cite{soucy2020estimating, kraemer2020effect, bryant2020estimating, gao2020mapping, chinazzi2020effect} exploring the interplay between human mobility responses, social distancing policies, and transmission dynamics in response to the COVID-19 pandemic. For example, it was shown by \cite{kraemer2020effect} that strict implementation of social distancing policies can reduce mobility and substantially mitigate the spread of COVID-19. A US mobility change map was created in \cite{gao2020mapping} to increase risk awareness of the public and visualize dynamic changes in mobility as COVID-19 situation and policy evolves. Study \cite{chinazzi2020effect} measures the effectiveness of non-pharmaceutical interventions (NPIs) introduced by
% governments across Europe using the changes of mobility.   Studies \cite{hellewell2020feasibility, cho2020contact} have also explored the feasibility of utilizing contact tracing to control the spread of the disease through simulated synthetic data and real-world smartphone trajectories. These studies are timely in showing the important role played by mobility in the spread of COVID-19, but they do not address the challenges in real-world mobility estimation/simulation(e.g., effects of unknown, uncertain, and random factors), and they analyze the mobility changes in city or country scale. Besides, these studies have not explored the potential use of deep learning based generative models to assist the estimation.

% \par \noindent
\textbf{Deep Learning for Spatio-Temporal Prediction.} There have been many deep learning based techniques developed for spatiotemporal data. For example, LSTM were widely used in traffic accident prediction \cite{yuan2018hetero}
%and flow prediction \cite{pan2019matrix},
due to its capability in capturing spatio-temporal correlation and thus provide good prediction results. %Geospatial object mapping \cite{timber}, 
Geospatial object mapping \cite{timber}, taxi driver behavior imitation \cite{zhang2019unveiling}, taxi demand \cite{zhang2021mlrnn}, travel time estimation \cite{xu2020mtlm}, etc, they all combine the deep learning model with spatio-temporal perspective in their model design and obtain good performance. Most of these techniques typically are stationary predictors (i.e., same result from two runs on same data) rather than generative models, and they do not consider the unknown factors in prediction, and their performance relies on large data sets. Besides, they do not leverage domain knowledge based constraints to assist learning (e.g., cGAN \cite{10.1145/3397536.3422261, zhang2019trafficgan}). 
%In addition, generative neural networks have not been explored to assist human mobility response estimation at fine scale in this COVID-19 pandemic.
% xie2018unsupervised
% xie2020locally, 

% \par \noindent
% \textbf{Generative Adversarial Networks (GAN)}. GANs were proposed by \cite{goodfellow2014generative}, and has achieved great performance in image generation domain, including image-to-image translation \cite{isola2017image}, image super-resolution \cite{ledig2017photo}, and text-to-image synthesis \cite{reed2016generative}. %Despite the success, an critical issue for GANs is known to be the unstable and sensitive to the choices of hyper-parameters in learning process. Several works have attempted to address the GANs training problem and improved the stability by designing new network architectures \cite{karras2019style}, modifying the learning objectives and dynamics \cite{mao2017least}, adding regularization methods to obtain stable gradients \cite{che2016mode}. 
% Beside image generation, recently, deep graph generative adversarial structure has been developed based on the concept of unsupervised learning. Existing architectures build upon generative model including GraphVAE \cite{simonovsky2018graphvae} and GraphGAN \cite{wang2017graphgan}, and they achieved good performance. However, generative model has not been applied on estimating human mobility problem as well as other human related movement research. 

\textbf{Meta-Learning.} Meta-learning learns new tasks quickly and effectively with a few examples. Existing optimization-based meta-learning algorithms such as MAML \cite{finn2018probabilistic} and Reptile \cite{nichol2018reptile} rely on optimization through gradient descent, and both are compatible with any model. %MAML produces a good initialization toward a new task with a few steps of gradient updates, and Reptile is iteratively trained on a sampled task by multiple gradient steps.   
Recently, the idea of optimization-based meta-learning has been applied to many domains including classification and reinforcement learning. However, there only are a few work address the spatial and temporal problems simultaneously. In traffic prediction, a recent work \cite{zhang2020cst} focuses on knowledge transfer in a single city, which only deals with temporal tasks with no spatial-based tasks. \cite{wang2018cross} proposes a transfer learning framework for traffic prediction through learning region matching function. Another work \cite{yao2019learning} which is based on multiple cities does not consider temporal patterns and dynamic scenarios. 
% This model is designed with no time-based tasks, which is insufficient to model the continued and dynamic changes.

\textbf{Mobility Estimation.} There have been many studies \cite{kraemer2020effect, gao2020mapping, chinazzi2020effect} exploring the interplay between human mobility responses, social distancing policies, and transmission dynamics in response to the COVID-19 pandemic. 
% For example, it was shown by \cite{kraemer2020effect} that strict implementation of social distancing policies can reduce mobility and substantially mitigate the spread of COVID-19. 
A US mobility change map was created in \cite{gao2020mapping} to increase risk awareness of the public and to visualize dynamic changes in mobility as COVID-19 situation and policy evolves. 
% Study \cite{chinazzi2020effect} measures the effectiveness of non-pharmaceutical interventions (NPIs) introduced by
% governments across Europe using the changes of mobility.   %Studies \cite{hellewell2020feasibility, cho2020contact} have also explored the feasibility of utilizing contact tracing to control the spread of the disease through simulated synthetic data and real-world smartphone trajectories. 
These studies are timely in showing the important role played by mobility in the spread of COVID-19, but they do not address the challenges in real-world mobility estimation/simulation (e.g., effects of unknown, uncertain, and random factors), and they analyze the mobility changes in city or country scale. A study \cite{chang2021supporting} simulated the human mobility which allows policymakers to inspect mobility changes under different policies. But this approach utilizes a traditional epidemiological model, and does not transfer the simulation from city to city by shared knowledge. 
%Another study \cite{jiang2021transfer} proposed a deep neural network model to capture spatio-temporal information from human mobility data through a straight forward parameter sharing method, and transferred from one city to another. 
However, these studies have not explored the potential use of deep learning based generative models and meta-learning to assist the estimation.

\section{Conclusions}
We made the first attempt to tackle the human mobility estimation problem through a spatio-temporal meta-generative framework. Specifically, we proposed a STORM-GAN model to capture complex spatio-temporal patterns using a set of social and policy conditions related to COVID-19. We also proposed a novel spatio-temporal task-based graph (STTG) to represent the spatio-temporal relationships among cities, with a graph convolution network to learn embeddings of its subgraph for cross-task learning enhancements. Finally, STORM-GAN utilized the meta-learning paradigm to learn shared-knowledge from a spatio-temporal distribution of estimation tasks and can quickly adapt to new tasks (e.g., new cities). The experiment results showed that our proposed approach can significantly improve the estimation performance compared to baselines. 
% The improvement is particularly crucial for policymakers. 
The model can assist policymakers to better understand
% By understanding 
the dynamic mobility pattern changes under different social and policy conditions, and can potentially be leveraged to inform decisions in resource allocation and provisioning, event planning, response management, etc. 

% As the estimation objective is not limited to mobility responses, users may also utilize the model to estimate other indicators (e.g., the number of cases) with different input conditions.
% try different mobility response measures or utilize the model structure to estimate pandemic cases or other related indicators using mobility responses as a condition.

% In future work, we will explore the consideration of fairness in the estimation for regions with different demographics. Furthermore, we plan to develop new formulations to address potential issues related to data quality and uncertainty in this problem.

\section*{Acknowledgment}
This paper is funded in part by Safety Research using Simulation University Transportation Center (SAFER-SIM). SAFER-SIM is funded by a grant from the U.S. Department of Transportation’s University Transportation Centers Program (69A3551747131). However, the U.S. Government assumes no liability for the contents or use thereof. Yiqun Xie is supported in part by NSF grants 2105133, 2126474, 2147195, Google's AI for Social Good Impact Scholars program, and the DRI award at the University of Maryland; and Xiaowei Jia is supported in part by NSF award 2147195, USGS award G21AC10207, Pitt Momentum Funds award, and CRC at the University of Pittsburgh. Yanhua Li was supported in part by NSF grants IIS-1942680 (CAREER), CNS-1952085, CMMI-1831140, and DGE-2021871. 

% We thank SafeGraph Inc. (www.safegraph.com) for providing free access to Boston POI data, including Core Places, Geometry, and Weekly Places Patterns for this research.

\bibliographystyle{IEEEtran}
\bibliography{sample-base}

\end{document}